\documentclass[acmtog,nonacm]{acmart}
\makeatletter
\def\@ACM@checkaffil{% Only warnings
    \if@ACM@instpresent\else
    \ClassWarningNoLine{\@classname}{No institution present for an affiliation}%
    \fi
    \if@ACM@citypresent\else
    \ClassWarningNoLine{\@classname}{No city present for an affiliation}%
    \fi
    \if@ACM@countrypresent\else
        \ClassWarningNoLine{\@classname}{No country present for an affiliation}%
    \fi
}
\makeatother
\acmSubmissionID{2309}

\usepackage{booktabs} % For formal tables
\usepackage{ulem}
\usepackage[ruled]{algorithm2e} % For algorithms

\SetAlFnt{\small}
\SetAlCapFnt{\small}
\SetAlCapNameFnt{\small}
\SetAlCapHSkip{0pt}

\renewcommand\footnotetextcopyrightpermission[1]{}
\begin{teaserfigure}
    \centering
    \includegraphics[width=\linewidth]{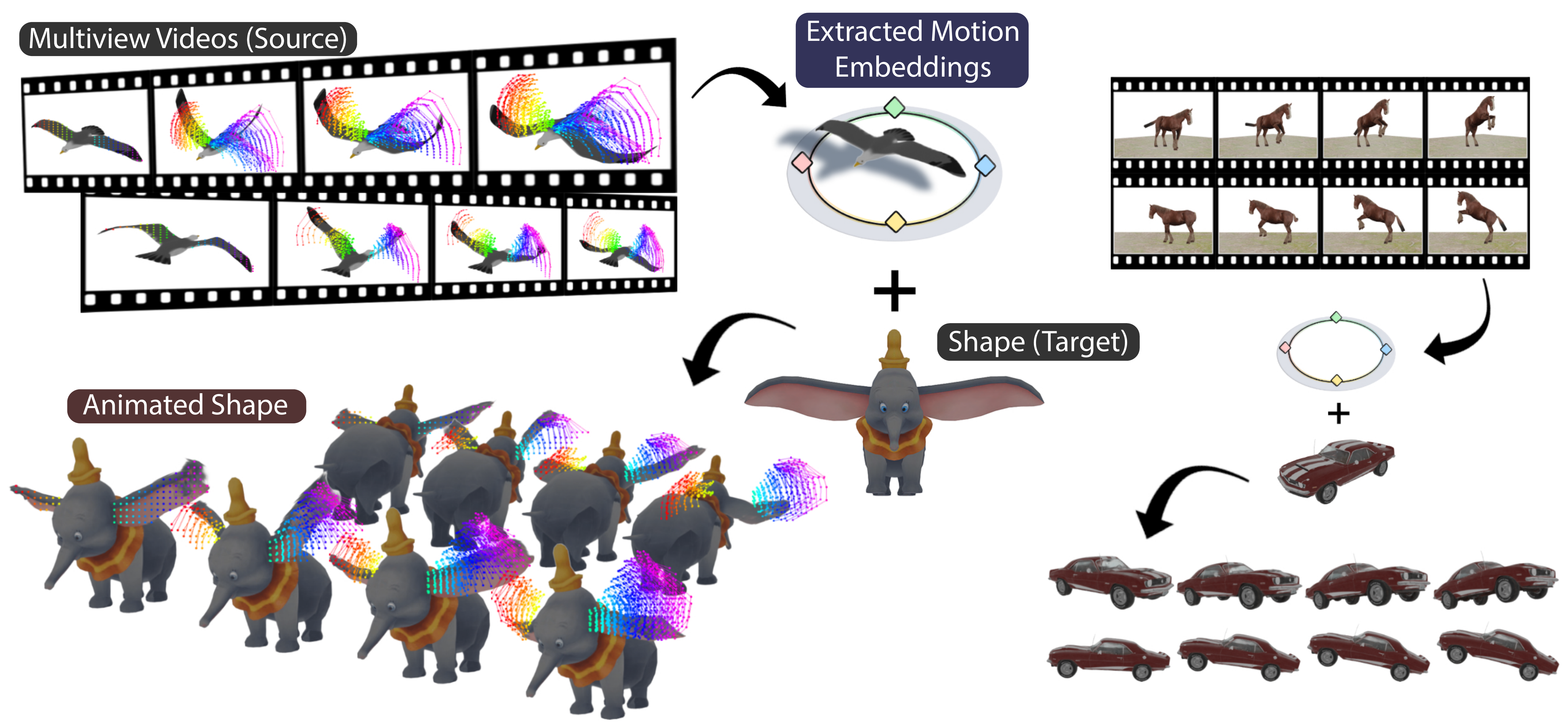}
    \caption{\textbf{Semantic 3D Motion Transfer in Action}. Our method extracts motion embeddings from a multiview video and applies them to a static 3D Gaussian Splatting (3DGS) asset, bringing it to life with motion that matches the semantics of the source. \textbf{Left:} A bird’s wing flapping motion is transferred to an elephant cartoon’s ears. \textbf{Right:} A horse’s rearing motion animates a vehicle lifting its front wheels. \textit{We encourage watching the supplementary video for a clearer depiction of motion, which is best appreciated in dynamic form.}}
    \label{fig:teaser}
\end{teaserfigure}

\begin{document}
% Title portion
\title{Gaussian See, Gaussian Do: Semantic 3D Motion Transfer from Multiview Video}

% DO NOT ENTER AUTHOR INFORMATION FOR ANONYMOUS TECHNICAL PAPER SUBMISSIONS TO SIGGRAPH 2019!
% \author{}
% \orcid{1234-5678-9012-3456}
% \affiliation{%
 % \institution{}
 % \streetaddress{104 Jamestown Rd}
 % \city{Williamsburg}
 % \state{VA}
 % \postcode{23185}
 % \country{USA}
 % }
\author{Yarin Bekor}
\authornote{Equal contribution.}
\affiliation{%
 \institution{Technion - Israel Institute of Technology}}
\email{yarin.bekor@campus.technion.ac.il}

\author{Gal Michael Harari}
\authornotemark[1]
\affiliation{%
 \institution{Technion - Israel Institute of Technology}}
\email{Gal.harari11@gmail.com}

\setcounter{footnote}{0}
\author{Or Perel}
\affiliation{%
 \institution{NVIDIA}}
\affiliation{%
 \institution{University of Toronto}}
\affiliation{%
 \institution{Vector Institute}}
\email{orr.perel@gmail.com}

\author{Or Litany}
\affiliation{%
 \institution{Technion - Israel Institute of Technology}}
\affiliation{%
 \institution{NVIDIA}}
\email{or.litany@gmail.com}

\begin{abstract}
We present Gaussian See, Gaussian Do, a novel approach for semantic 3D motion transfer from multiview video. Our method enables rig-free, cross-category motion transfer between objects with semantically meaningful correspondence. Building on implicit motion transfer techniques, we extract motion embeddings from source videos via condition inversion, apply them to rendered frames of static target shapes, and use the resulting videos to supervise dynamic 3D Gaussian Splatting reconstruction. Our approach introduces an anchor-based view-aware motion embedding mechanism, ensuring cross-view consistency and accelerating convergence, along with a robust 4D reconstruction pipeline that consolidates noisy supervision videos. We establish the first benchmark for semantic 3D motion transfer and demonstrate superior motion fidelity and structural consistency compared to adapted baselines. Code and data for this paper available at \href{https://gsgd-motiontransfer.github.io/}{gsgd-motiontransfer.github.io}.
\end{abstract}



\maketitle

\section{Introduction}
\label{sec:intro}

Animation is the art of bringing still objects and characters to life. The demand for realistic 3D animation is rapidly growing across industries, from gaming and virtual/augmented reality (VR/AR) to robotics and autonomous system simulations. As 3D content reconstruction and generation scales, the need for a controllable, data-driven approach to animating 3D objects across diverse categories is more important than ever. However, generating high-quality 3D motion remains a significant challenge. 
A major bottleneck is the lack of methods that enable 3D objects---especially those without predefined kinematic structures---to acquire motion in a semantically meaningful way. Traditional motion generation relies heavily on rigging, where a predefined skeletal structure governs movement. While well-established techniques exist for animating humans and, to some extent, animals~\cite{baran2007automatic, ma2023tarig, li2021learning, xu2020rignet}, applying these principles to arbitrary, non-rigged objects remains an open challenge~\cite{liu2025riganything, chu2024humanrig}. Recent advances in 3D Gaussian Splatting (3DGS)~\cite{guo2024make} have streamlined the reconstruction of 3D assets from images, yet these representations lack an inherent motion structure, making animation even more difficult.

Even if we could control such objects, instructing them to move meaningfully is another challenge. Some methods rely on textual descriptions of motion~\cite{tevet2022human, zhang2024motiondiffuse, petrovich2022temos,singer2023text4d,ling2024alignyourgaussians,bah20244dfy,ren2023dreamgaussian4d,wang2024animatabledreamer,zhao2023animate124,zheng2024unified}, but accurately describing complex motion in words is inherently limited~\cite{petrovich2024multi}. Motion is often too nuanced for text-based instructions to capture effectively, making such methods imprecise and difficult to generalize. A more natural approach is motion transfer---enabling an object to mimic motion demonstrated by another. While much research has focused on motion retargeting within the same category~\cite{sun2022human, raab2024monkey, aberman2020skeleton, chen2023weakly, li2021learning, villegas2018neural, zhang2023skinned}, transferring motion between objects that do not share an explicit kinematic mapping is significantly more challenging~\cite{jacobianrigfree}. 

Recently, semantic motion transfer techniques have emerged in 2D video generation~\cite{kansy2024reenact, xiao2024video, ling2024motionclone, wang2024motioninversionvideocustomization, wei2024dreamvideo, zhao2023motiondirector, materzynska2023customizing}, disentangling motion intent from appearance and transferring it to different targets. However, these capabilities remain absent in 3D, despite their significant potential and necessity.

We propose a novel method for semantic 3D motion transfer from multiview video, allowing motion from a dynamic source object to be adapted to a static 3D target object while ensuring realistic and coherent movement. 

We assume access to multiview video of the source motion to recover its full dynamics, since a single-view input can miss occluded regions—for example, an octopus articulating its legs differently. Such multiview videos may come from either computer animation, or multiview captures obtainable with simple off-the-shelf equipment, as demonstrated by the AIST Dance dataset \cite{aist-dance-db}.

Uniquely, our method does not rely on structural similarity between the source and target, enabling motion transfer across vastly different object categories---such as a horse rearing to a sports car lifting its front wheels. Building on condition inversion in generative video models~\cite{gal2022image,kansy2024reenact}, we extend these principles to 3D through a view-aware motion embedding mechanism. Specifically, we introduce an anchor-based interpolator that optimizes embeddings collaboratively across views, significantly accelerating convergence while enhancing motion fidelity. We use these embeddings to generate target supervision videos, which, while capturing the intended motion, often contain artifacts and inconsistencies.

To address these imperfections, we propose a robust 4D reconstruction pipeline that effectively transforms noisy supervision signals into high-quality, temporally stable motion reconstructions. Specifically, we employ a 3D Gaussian splatting representation of the target object and infer its animation by optimizing a motion field applied to a set of control points governing deformation. Through carefully designed regularization strategies, we demonstrate that even when supervision videos are inconsistent, our method produces realistic and stable 4D reconstructions.

To facilitate rigorous evaluation and drive progress on this underexplored and challenging problem setup, we introduce the first benchmark for semantic 3D motion transfer from multiview video. It spans diverse, cross-category motion scenarios and includes carefully curated source-target pairs of structurally different 3D objects, along with tailored evaluation metrics for both motion fidelity and structural consistency.

\noindent Our work makes the following key contributions:
\begin{itemize} 
\item We introduce a \textbf{new problem setup for semantic 3D motion transfer}, where motion from a multiview video of a dynamic source object is adapted to a static 3D Gaussian Splatting (3DGS) target, even across vastly different categories, without requiring predefined skeletal correspondences or structural similarity.

\item We propose a novel \textbf{view-aware motion embedding} strategy, optimized collaboratively through an anchor-based interpolator. This approach balances global coherence with view-dependent details, ensuring motion fidelity while significantly accelerating convergence.

\item We develop a \textbf{robust 4D reconstruction pipeline} that refines noisy, artifact-prone supervision videos into high-quality, temporally stable dynamic 3D reconstructions.

\item We introduce a \textbf{benchmark for semantic 3D motion transfer}, evaluating diverse cross-category motion transfers between structurally different 3D objects.
\end{itemize}

Our method outperforms adapted baselines in both motion fidelity and structural consistency, delivering high-quality motion transfer across diverse, cross-category scenarios. A user study further confirms that our results are consistently preferred by human evaluators. Beyond synthetic settings, we also demonstrate compelling results on real-world scenes by animating 3D assets reconstructed from in-the-wild imagery (see Fig.~\ref{fig:inthewild}), highlighting the practicality and robustness of our approach.

\section{Related Work}
\label{sec:relatedwork}

\paragraph{4D Generation.} 
Over the past years, there has been a surge of research in 4D content generation. \cite{wu2024sc4d, miao2024pla4d, zeng2024stag4dspatialtemporalanchoredgenerative, ren2023dreamgaussian4d, jiang2024consistentd, uzolas2024motiondreamer, bah20244dfy, bah2024tc4d, ling2024alignyourgaussians} primarily focus on full generation, where both the 3D target object and its motion are synthesized from textual input. Despite their capabilities, these generation methods cannot incorporate unique motion patterns from inputs videos, or to apply movement while preserving the target's distinctive identity.
\cite{ling2024alignyourgaussians, ren2023dreamgaussian4d} split their pipeline, by first generating a static 3D figure, then animating it, potentially allowing for target identity preservation.

\paragraph{3D Motion Transfer.} 
Transferring motion from source input onto a target 3D figure, often referred to as "Motion Retargeting", is a well-studied problem \cite{raab2024monkey, zhang2024generative, villegas2018neural, aberman2020skeleton, zhang2023skinned, wang2025towards, kim2024most}. 
However, these methods typically rely on rigged figures and are predominantly constrained to human characters. In contrast, our approach allows for motion transfer onto arbitrary 3D objects without requiring manual rigging, also enabling cross-category motion transfer. 
One exception is the generative work of SC4D \cite{wu2024sc4d} which introduces 3D motion transfer without requiring a rigged figure.

\paragraph{Generative Models Based Motion Transfer.}
Recent years have witnessed significant advancements in video generation models \cite{blattmann2023stable, guo2024animatediffanimatepersonalizedtexttoimage, xing2023dynamicrafter, zhang2023i2vgenxlhighqualityimagetovideosynthesis, wang2023modelscopetexttovideotechnicalreport}. These models provide a rich motion prior, encouraging researchers to distill, represent, and apply the motion learned by these models~\cite{kansy2024reenact, yatim2024space, xiao2024video, ling2024motionclone, wang2024motioninversionvideocustomization, wei2024dreamvideo, zhao2023motiondirector, materzynska2023customizing}. Typically, these approaches extract motion from a reference video and apply it as a motion embedding to a new image or prompt, generating a new video that replicates the reference motion. However, they remain limited to 2D motion transfer.
Our approach extends this paradigm to the 3D domain, leveraging the 2D learned motion priors to enable rigless, cross-category motion transfer, effectively bridging the gap between 2D motion learning and 3D animation.
\section{Preliminaries}
\label{sec:preleminaries}

\textbf{Semantic 2D Motion Transfer.}  Our setting assumes a conditional latent Image-to-Video diffusion model  \cite{blattmann2023stable}, which also uses an additional embedding input, e.g. a CLIP \cite{Radford2021LearningTV} embedding of a text prompt or image.
\cite{kansy2024reenact} demonstrated that such video diffusion models \cite{blattmann2023stable, zhang2023i2vgenxlhighqualityimagetovideosynthesis}
exhibit an implicit disentanglement in the way motion and appearance are encoded, where the image input controls the \textit{appearance} and the embedding input manipulates the \textit{motion} of the generated video.

Based on this finding, Kansy et al. \cite{kansy2024reenact} use condition embedding inversion \cite{gal2022image} to learn a multiframe representation $\mathbf{m}^* \in \mathbb{R}^{M \times (F+1) \times d}$, where $M$ denotes the number of $d$-dimensional tokens, capturing the motion of a source video $\mathbf{x}_0 \in \mathbb{R}^{F \times H \times W \times 3}$ with $F$ frames of size $H \times W$. This motion representation can then be transferred to animate images of different subjects while largely preserving the semantic intent of the original motion.

Formally, the process is optimized through a frozen video diffusion denoiser $D_\theta$ by minimizing the denoising score matching loss:

% \vspace{-0.4cm}
\begin{equation}
\mathbf{m^*} = \underset{\mathbf{m}}{\mathrm{arg\,min}} \; 
\mathbb{E}
\left[ \lambda_\sigma \|D_\theta(\mathbf{x}_0 + \boldsymbol{n}; \sigma, \boldsymbol{m}, \boldsymbol{c}) - \mathbf{x}_0\|_2^2\right],
\end{equation}
% \vspace{-0.4cm}

\noindent where the expectation is w.r.t $\sigma,\boldsymbol{n} \sim p(\sigma,\boldsymbol{n})$, $\boldsymbol{c}$ encompasses condition signals (e.g. first frame), and \\$p(\sigma,\boldsymbol{n})=p(\sigma)\mathcal{N}(\mathbf{n}; \mathbf{0}, \sigma^2\boldsymbol{I})$ with $p(\sigma)$ being the probability distribution over noise level $\sigma$. Then $\boldsymbol{n}$ denotes the noise, and $\lambda_\sigma$ is a weighting function. For the rest of the manuscript we will drop $\boldsymbol{c}$ and replace $\mathbf{x}_0 + \boldsymbol{n}$ with $\widetilde{\mathbf{x}}_0$.

\label{sec:scgs}
\noindent \textbf{3D Gaussian Splatting and Low-Dimension Controllability.} Our setting assumes an input static target object, represented with 3D Gaussian Splats (3DGS) \cite{kerbl20233d}. 3DGS  optimizes differentiable primitives to model volumetric radiance fields. Specifically, Gaussian particles are projected and alpha-composited through rasterization to obtain high-quality 3D reconstructions of objects and scenes. An object can be represented as a set of 3D Gaussians $\mathcal{G}_i$, each parameterized with $\bigl\{ \boldsymbol{\mu}_i, \boldsymbol{q}_i, \boldsymbol{s}_i, \sigma_i, \boldsymbol{c}_i \bigr\}$, with mean $\boldsymbol{\mu}_i$, rotation $\boldsymbol{q}_i$, scale $\boldsymbol{s}_i$, opacity $\sigma_i$ and view dependent color $\boldsymbol{c}_i$ modeled with Spherical Harmonic coefficients. The contribution of a single Gaussian $\mathcal{G}_i$ to the radiance field at point $\boldsymbol{x} \in \mathbb{R}^3$ is expressed by the probability density function: \( \mathcal{G}(\boldsymbol{x}) = e^{-\frac{1}{2}(\boldsymbol{x}-\boldsymbol{\mu}_i)^T \Sigma_i^{-1}(\boldsymbol{x}-\boldsymbol{\mu}_i)}\), where the covariance can be decomposed to \(\Sigma_i = R_iS_i S_i^T R_i^T\), with $R_i, S_i$ being the rotation matrix derived from the rotation quaternion $\boldsymbol{q}_i$ and scale vector $\boldsymbol{s}_i$ respectively. The final color of a pixel is determined by all Gaussians that overlap it, by alpha-blending in depth order: 

% \vspace{-0.3cm}
\begin{equation}
    \mathcal{C} = \sum_i \boldsymbol{c}_i \alpha_i \prod_{j=1}^{i-1} (1 - \alpha_j)  
\end{equation}
% \vspace{-0.2cm}

\noindent Here, $\alpha_i$ is a function of each Gaussian's opacity and PDF post-projection to image coordinates. 

Building on this foundation, SC-GS \cite{huang2024sc} extends 3DGS for dynamic scenes by 
decomposing canonical appearance and motion control using 3D Gaussians $\mathcal{G}_i$ and control points 
$\bigl\{ \boldsymbol{p}_k \in \mathbb{R}^3 \bigl\}$ with a learnable coordinate and radius. A time-conditioned MLP predicts the rotation and translation of control points at time $t$ as: $\Psi: (\boldsymbol{p}_k, t) \rightarrow (R_k^t|T_k^t)$.
Gaussians are then deformed using Linear Blend Skinning (LBS) \cite{sumner2007embedded}, where the warped position \(\boldsymbol{\mu}_i^t\) and rotation \(\boldsymbol{q}_i^t\) for Gaussian \(\mathcal{G}_i\) are computed as:

% \vspace{-0.2cm}
\begin{equation}
% \begin{split}
    \boldsymbol{\mu}_i^t = \sum_{k \in \mathcal{N}_i} w_{ik} \Bigl( R_k^t(\boldsymbol{\mu}_i - \boldsymbol{p}_k) + \boldsymbol{p}_k + T_k^t \Bigl) ;
    \boldsymbol{q}_i^t = \Bigl( \sum_{k \in \mathcal{N}_i} w_{ik} \boldsymbol{r}_k^t \Bigl) \otimes \boldsymbol{q}_i
% \end{split}
\label{eq:scgs_skinning}
\end{equation}
% \vspace{-0.2cm}

\noindent such that the weights \(w_{ik}\) are based on the distances \(d_{ik}\) between the center of Gaussian \(\mathcal{G}_i\) and its neighboring control points \(\mathcal{N}_i \) and their learned radii. $\boldsymbol{r}_k$ is the quaternion form of $R_k$ and $\otimes$ symbolizes a product of quaternions.

% \section{Method}
\section{Semantic 3D Motion Transfer}
\label{sec:method}
\begin{figure*}[h]
    \centering
    \includegraphics[width=\linewidth]{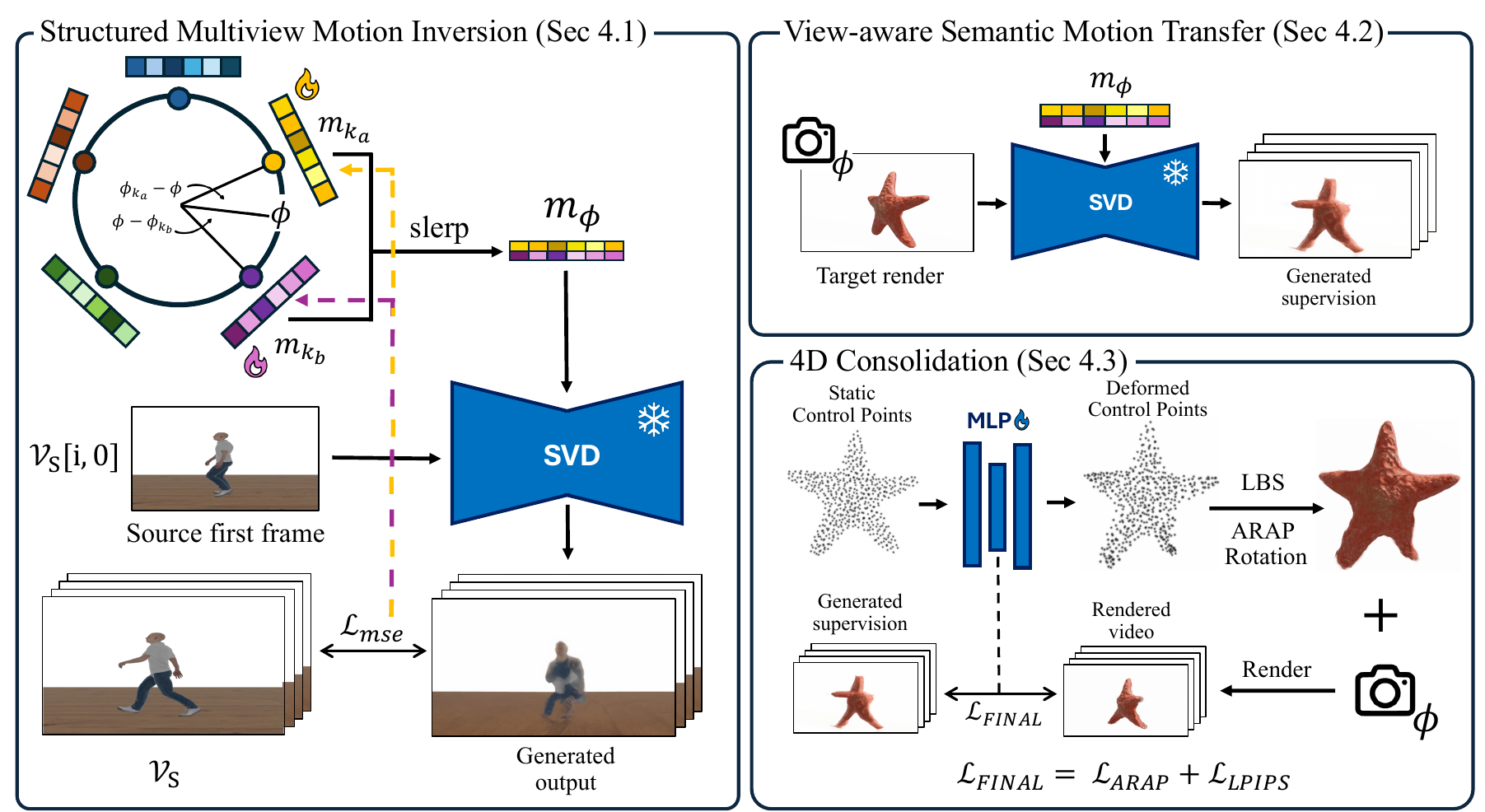}
    \caption{Pipeline Overview. (1) \textbf{Structured Multiview Motion Inversion.} We extract motion embeddings from the source using the slerp interpolation from the two nearest achor points. (2) \textbf{View-aware Semantic Motion Transfer.} We use the motion embeddings to generate supervision for the motion transfer process, and then (3) \textbf{4D Consolidation.} We apply the  supervision onto the target shape to introduce actual motion}
    \label{fig:Method}
\end{figure*}

We aim to transfer the dynamics of a source object $\mathcal{V}_S$, represented through a sparse set of $N$ videos captured by static cameras positioned at different views, to a static target object of a different structure and posture, $\mathcal{G}_T$, represented as 3DGS~\cite{kerbl20233d} (Section~\ref{sec:scgs}). Our goal is to reenact $\mathcal{G}_T$ with the semantic motion patterns of $\mathcal{V}_S$ while preserving its identity, appearance and geometry as faithfully as possible.

Formally, we take as input tensor $\mathcal{V}_S \in \mathbb{R}^{N \times F \times H \times W \times 3}$, of $N$ videos of $F$ frames, where each video is associated with known camera extrinsics $[R_{C_{i}}|T_{C_{i}}] \in SE(3)$. Our pipeline is made of two stages. First, we extract structured motion embeddings $\boldsymbol{m}_1^*..\boldsymbol{m}_K^*$ from $\mathcal{V}_S$ using a pre-trained video diffusion model $D_\theta$ \cite{blattmann2023stable}, where $K \leq N$. In the second stage, we fix a set of control points $\mathcal{P}_T$, and use them to drive the target shape $\mathcal{G}_T$. This process is governed by optimizing a decoder $\Psi$, with self-supervised labels generated by frozen video diffuser $D_\theta$ conditioned on the previously extracted motion embeddings $\boldsymbol{m}_1^*..\boldsymbol{m}_K^*$. Our setting allows the same motion code to be reused over multiple target shapes.  

\subsection{Structured Multiview Motion Inversion}
\label{sec:Structured_Multiview}

Inspired by the 2D motion inversion introduced in \cite{kansy2024reenact}, we devise a motion inversion process in 3D by extracting motion embeddings from each input view-video. Lifting the inversion technique of \cite{kansy2024reenact} to 3D, however, is not straightforward. Through experimentation, we observe that trained motion embeddings that have been optimized for a specific angle are not necessarily suitable for generating motion from a different angle, as shown in Figure~\ref{fig:Novel_View}. Intuitively, one can imagine that the pixels depicting a "kicking" animation appearing somewhat different in orthogonal directions. One simple solution is to optimize motion embeddings \( \boldsymbol{m}_1^*, \dots, \boldsymbol{m}_{N}^* \) for each source view $\{V_S[i]\}_{i=1}^{N}$:

% \vspace{-0.4cm}
\begin{equation}
\label{eq:simple_embeddings}
\boldsymbol{m}_i^* = \underset{\boldsymbol{m}_i}{\mathrm{arg\,min}} \; \mathbb{E}
\left[ \lambda_\sigma \|D_\theta(\widetilde{\mathcal{V}}_S[i]; \sigma, \boldsymbol{m}_i) - \mathcal{V}_S[i]\|_2^2\right],
\end{equation}
% \vspace{-0.4cm}

\noindent and use them along with corresponding rendered views of the static 3DGS target object to generate \( N \) videos of the target object at matching angles, which then serve as supervision for 4D reconstruction. However, this approach is suboptimal for several reasons. First, it is inefficient in terms of runtime and storage costs - as it requires training and bookkeeping a separate motion embedding for each possible source angle. That leads to increased computational and storage costs as the number of embeddings grows linearly with the number of videos in $\mathcal{V}_S$.
Additionally, this method lacks generalization, as embeddings trained on specific angles cannot be reused or adapted to unseen viewpoints, further limiting its flexibility and applicability.

To overcome these limitations, we hypothesize that since motions in different views share a common underlying 3D motion, their embeddings should share information. Since that does not seem to naturally occur in the standard per view inversion procedure, we propose to instill this relation by inducing view structure into the optimization procedure. By doing so we aim to not only make the resulting embeddings more generalizable but also speed up the optimization procedure. Specifically, we propose to optimize a fixed number of $K$ anchor motion embeddings \(\{\boldsymbol{m}_i\}_{i=1}^{K}\), distributed evenly across a range of source view angles \( \{ \phi_i \}_{i=1}^K \). When a motion embedding for a specific angle $\phi \in \mathbb{R}$ is required, we perform a spherical linear interpolation between the two closest anchor embeddings with:

% \vspace{-0.5cm}
% \vspace{-0.2cm}
\begin{equation}
\begin{aligned}
 & \boldsymbol m_\phi = \text{slerp}(\boldsymbol{m}_i, \boldsymbol{m}_j, \frac{\phi-\phi_i}{\phi_j - \phi_i}) ~~\text{s.t.} \\ &i = \underset{k}{\arg\max} \{\phi_k | \phi_k \leq \phi\} ~ ; ~ j = {(i \bmod K) + 1} 
 \label{eq:interpolation}
\end{aligned}
\end{equation}
% \vspace{-0.2cm}
% \vspace{-0.4cm}

Here, $\text{slerp}(\cdot, \cdot; \phi)$ denotes Spherical Linear Interpolation, ensuring smooth transitions between the closest anchor embeddings $\boldsymbol{m}_i^*$ and $\boldsymbol{m}_j^*$. The parameter $\phi$ controls the interpolation, allowing for a continuous blend between the two anchors. This formulation enables the model to dynamically adjust motion embeddings while maintaining efficiency and scalability. To sum up, each iteration of our training loop samples a single video $\mathcal{V}_S[i]$ of angle $\phi$, and uses Equation~\ref{eq:interpolation} to obtain an interpolated code $m_\phi$, which we then feed into the score matching function:

% \vspace{-0.2cm}
\begin{multline}
\footnotesize
\{\boldsymbol{m}_k^*\}_{1}^K = \underset{\{\boldsymbol{m}_k\}}{\mathrm{arg\,min}} \; \mathbb{E} \left[ \lambda_\sigma \|D_\theta(\widetilde{\mathcal{V}}_S[i]; \sigma, \boldsymbol{m}_\phi) - \mathcal{V}_S[i]\|_2^2\right].
\label{eq:gsgd_denoise}
\end{multline}
% \vspace{-0.2cm}

It then follows that our backwards pass updates multiple anchor motion embeddings at once.

\subsection{View-aware Semantic Motion Transfer}
After extracting motion embeddings \(\{\boldsymbol{m}_i^*\}_{i=1}^{K}\) from the multi-view video $\mathcal{V}_S$, our pipeline can readily apply them to a 3D shape represented with 3DGS \cite{kerbl20233d} for flexible motion transfer. First, we render an image of the target, $\mathcal{R}[\mathcal{G}_T ; C]$, with random camera $C \in SE(3)$, sampled uniformly. Then, we use it to generate 2D supervision video $\mathcal{V}_T$ by conditioning the denoising process at the right side of Equation~\ref{eq:gsgd_denoise} on the rendered image $\mathcal{R}[\mathcal{G}_T ; C]$, and its corresponding motion embedding $\boldsymbol{m}_\phi$. This process can be applied to any number of views, as motion embeddings can be interpolated between anchors, allowing flexible generation of supervision videos from different perspectives.

\subsection{4D Consolidation}
\label{sec:consolidation}
Our final step is a consolidation process, transforming the generated supervisions into a 4D representation. Since the static target shape is already provided, we model the 4D scene as a dynamic-3DGS representation leveraging the SC-GS foundation (Section~\ref{sec:scgs}), e.g. our goal is to drive the 3DGS static shape with control points. We begin by initializing a fixed set of control points $\mathcal{P}_T$ over target shape $\mathcal{G}_T$, using furthest point sampling. Then, we use the skinning scheme depicted in Equation~\ref{eq:scgs_skinning} with a relaxed weighting scheme: \(w_{ik} = exp \bigl( {-{d_{ik}}^2}/{2\beta^2} \bigl)\), 
\noindent where $d_{ik}$ is the distance between gaussian $i$ and control point $k$, and $\beta$ is a non-learned scalar which modulates the area of effect per control point.

Given multi-view supervision videos $\mathcal{V}_T$, we optimize a deformation field network $\Psi$ to map each control point $\boldsymbol{p}_k \in \mathcal{P}_T$ to a transformation $(R_k^t|T_k^t)$ that when applied, aligns $\mathcal{G}_T$ with the animation depicted within the supervision  $\mathcal{V}_T$. That is, our loss function minimizes the discrepancy between the rendered dynamic 3DGS and the supervision videos. The supervision videos, being the output of a video generative model, are inherently noisy and spatially inconsistent, see Appendix \ref{app:noisy_supervisions}. Hence, using them directly as supervision leads to a poorly reconstructed object (Figure~\ref{fig:ablation_quality}, Vanilla).

\paragraph{ARAP Loss and ARAP Rotation.} The As-Rigid-As-Possible (ARAP) loss \cite{sorkine2007rigid} is a widely used technique \cite{luiten2023dynamic, huang2024sc, li2024dreammesh4d} for preserving local rigidity during deformation by minimizing distortions in the transformed shape. In its standard form, ARAP loss is computed for each control point $p_k \in P_T$ and its nearest neighbors $\mathcal{N}_k$, ensuring that local relative positions remain consistent across transformations: 
% \vspace{-0.1cm}
\begin{equation} 
    \mathcal{L}_{ARAP} = \sum_{t\in{1..F}}\sum_{i\in\mathcal{N}k } w_{ik}|| (p^{t}_i - p^{t}_k) - \hat{R}^t_k (p_i - p_k) ||^2 
\end{equation}
% \vspace{-0.15cm}
where the optimal rotation $\hat{R}^t_k$ is estimated as:
\begin{equation}
\label{eq:arap_rot}
\hat{R}^t_k = \underset{R}{\mathrm{argmin}} \sum_{i\in\mathcal{N}_k} w_{ik} || (p^t_i - p^t_k) - R (p_i - p_k) ||^2 
\end{equation}
% \vspace{-0.2cm}

However, a key limitation of the standard ARAP loss is that it only regularize the positions of control points while leaving their associated rotations $R^t_k$ unconstrained. This lack of rotational supervision leads to visible artifacts in the deformation process, as seen in Figure~\ref{fig:ablation_quality}, 'Vanilla'. These artifacts arise from inaccurate rotation predictions by the mlp $\Psi$ which result from noisy and inconsistent supervision. To address this issue, we propose an ARAP Rotation mechanism, which explicitly enforces rigid rotational constraints during training. Specifically, instead of relying on the MLP-predicted rotation $R^t_k$, we replace it with the optimized rotation $\hat{R}^t_k$ which is derived by minimizing the same rigid motion assumption as in Equation~\ref{eq:arap_rot}.
In contrast to previous approaches, where such explicit constraints are typically applied only in a post-processing step, we integrate this constraint directly into the training process.

\paragraph{Perceptual Loss.} When optimizing 4D reconstruction against consistent and clean videos, MSE and mask loss are a popular and pragmatic choice \cite{li2024dreammesh4d,wu2024sc4d}. However, in scenarios with spatial inconsistencies and significant noise, as is in our supervision videos, pixel-wise comparisons become insufficient, and the learned deformations tend to inherit the spatial inconsistency from the supervision itself, resulting in distorted reconstructed structure.\\
To address this, we replace pixel-wise losses with the LPIPS \cite{zhang2018perceptual} loss, which operates in a perceptual space rather than enforcing explicit pixel-wise correspondences. This approach has previously been shown to be effective previously \cite{wu2024reconfusion} in mitigating reconstruction artifacts.
By leveraging perceptual similarity rather than direct pixel alignment, our model becomes more robust to view inconsistencies, reducing unwanted deformations introduced by artifacts and leading to more stable and visually coherent reconstructions.

\paragraph{SDS Loss.} A widely used strategy for reconstruction with generative models under view inconsistencies is the Score Distillation Sampling (SDS) loss~\cite{poole2022dreamfusion}, commonly adopted in 4D generation (e.g. ~\cite{wu2024sc4d}). We experimented with adding either plain SDS Loss or Iterative Dataset Update \cite{instructnerf2023}, but found both to underperform compared to our method (see \ref{app:reconstruction-ablation}).

\section{Semantic Motion Transfer Benchmark}
\label{sec:dataset}

We introduce the first benchmark for general, cross-category 3D motion transfer from multiview video. Since no existing benchmark supports this task, we construct one to enable systematic evaluation. Our benchmark combines curated source videos and static 3D target objects drawn from a subset of the Mixamo dataset~\cite{Mixamo} and from publicly available assets collected from the web. Notably, the Mixamo subset includes paired motions—the same motion performed by both a source and a target character—allowing for reference-based evaluation, as described in the next section. The web-crawled, cross-category subset features a diverse range of motions (human, animal, and object) and targets (from skeletons to robotic arms), supporting broad evaluation of motion transfer across semantically and structurally different entities. All target assets were reconstructed using 3DGS. A detailed explanation about the benchmark can be found in Appendix~\ref{app:benchmark}.

\section{Experimental Evaluation}
\label{sec:experiments}
\begin{figure*}[h]
    \centering
    \centerline{
        \includegraphics[width=0.85\linewidth]{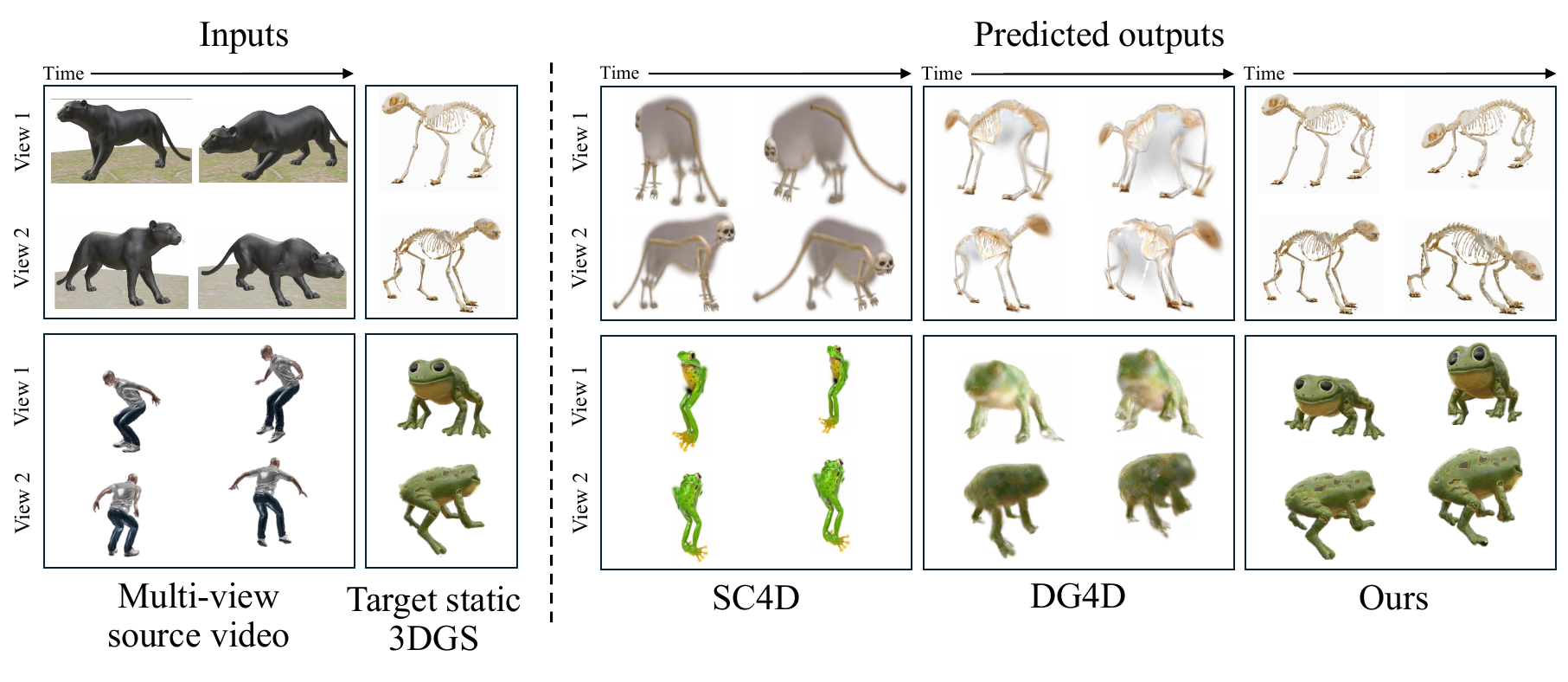}
    }
    \caption{\textbf{Qualitative comparison of semantic 3D motion transfer.} We compare our method to adapted baselines, showing two views of the source motion, the target 3DGS object, and the generated output for each method. We demonstrate superior identity preservation while also accurately transferring the source motion to the target. }
    \label{fig:qualitative_comparison}
\end{figure*}

% \caption{\textbf{Qualitative comparison of semantic 3D motion transfer.} Each row shows a target object and a source motion. The top and bottom rows are from the web-crawled dataset, while the middle row is from Mini-Mixamo. Our method achieves high-quality transfer across diverse and challenging categories.}

\paragraph{Evaluation Metrics.} As no established metric exists for semantic motion transfer in 3D, we adopt the \textit{Motion Fidelity} measure from \cite{yatim2024space}, originally proposed for 2D video motion. We also assess appearance preservation using \textit{CLIP-I} \cite{rahamim2024bringing,zhao2023animate124}, and similarity to ground truth via the \textit{CLIP similarity score}. See Appendix~\ref{app:eval_metric} for details.

% \subsection{Methods in comparison}
\paragraph{Methods in Comparison.}
To the best of our knowledge, no prior work tackles semantic motion transfer from multiview video to a 3DGS-represented target object. We therefore adapt the most relevant methods to our setting to enable meaningful comparisons.

Our first baseline is SC4D~\cite{wu2024sc4d}, a 2D-to-4D generation method that takes a 2D video and a textual prompt describing the target's appearance to synthesize a dynamic 3DGS output. Unlike our approach, SC4D does not perform direct motion transfer but instead \textit{modifies the appearance} of the generated object based on the prompt. To align it with our setup, we adapt SC4D in two ways: (1) we run its 2D-to-4D optimization using a primary view of the source video to capture relevant motion patterns; (2) we employ GPT-4o to generate a detailed textual description of the target, which SC4D uses to condition its appearance generation and match the intended object.

Our second baseline is DreamGaussians4D (DG4D)\cite{ren2023dreamgaussian4d}, an image-to-4D model that generates a dynamic 3DGS output from a single 2D image. DG4D first produces a driving video and a static 3D reconstruction, then optimizes a deformation field using Score Distillation Sampling (SDS) with a generative view synthesis model. To adapt it to our task, we replace the single input image with the full 3D target object and substitute the driving video with a reenacted source motion sequence, as described in Section\ref{sec:Structured_Multiview}. 

For our method, we use 5 anchor embeddings, 16 input views, 3K inversion steps, and 5K optimization iterations during 4D reconstruction. Further implementation details are provided in Appendix~\ref{app:method}.

\subsection{Comparing with Baselines}

% \begin{table*}[ht] % Use table* and [t] for top placement
%     \centering
%     \caption{Benchmark evaluation table}
%     \label{tab:main_experiment}
%     \begin{tabular}{lcccccc}
%         \toprule
%         & \multicolumn{2}{c}{\textbf{Motion Fidelity} $\uparrow$} & \multicolumn{3}{c}{\textbf{Pick-A-Pic}} & \textbf{CLIP Score} \\
%         \cmidrule(lr){2-3} \cmidrule(lr){4-6} \cmidrule(lr){7-7}
%         & Mini-Mixamo & Mini-Objaverse & Score (vs Prompt) & Score (vs GT) & Cosine (vs GT) & Mini-Mixamo \\
%         \midrule
%         SC4D Motion Transfer & 0.6539 & 0.5585 &  \textbf{19.2746} $\bullet$ & 96.807768 & 0.989598 & 0.9045 \\
%         2D Reenact + DreamGaussians4D  & 0.6115 & 0.5428 & 18.1906 & 97.644125 & 0.993828 & 0.9451 \\
%         % \midrule
%         % \textbf{Our Methods} \\
%         % \midrule
        
%         \textbf{Ours} & \textbf{0.7384} $\bullet$ & fill & 18.8066 & \textbf{98.202752} $\bullet$ & \textbf{0.996653} $\bullet$ & \textbf{0.9628} $\bullet$ \\
%         \midrule
        
%         GT full dataset & 0.874 & - & 18.6250 & 98.8645 & 1.0 & 1.0 \\
%         \bottomrule
%     \end{tabular}
% \end{table*}
% Motion Fidelity measures how faithfully the target shape follows the source motion. CLIP Score measures the semantic similarity of \yarinF{the ground-truth and predicted} rendered frames. Metrics are evaluated on the benchmarks described in Section~\ref{sec:Benchmark}.}

\begin{table}[ht] % Use table* and [t] for top placement
    \centering
    \caption{\textbf{Quantitative Comparison with Baselines.} Our method achieves significantly higher Motion Fidelity, CLIP-I, and CLIP similarity scores across both benchmarks compared to the baselines. We encourage the reader to refer to Figure~\ref{fig:qualitative_comparison} for a visual illustration of these improvements.}
    \label{tab:main_experiment}
    \resizebox{\columnwidth}{!}{
    \begin{tabular}{lcccccc}
        \toprule
        & \multicolumn{2}{c}{\textbf{Motion Fidelity} $\uparrow$} & \multicolumn{2}{c}{\textbf{CLIP-I} $\uparrow$} & \textbf{CLIP Score} $\uparrow$ \\
        \cmidrule(lr){2-3} \cmidrule(lr){4-5} \cmidrule(lr){6-6}
        & Mini-Mixamo & Cross-Category & Mini-Mixamo & Cross-Category & Mini-Mixamo \\
        \midrule
        SC4D Motion Transfer & 0.65 & 0.56 & 0.888 & 0.872 & 0.905 \\
        DreamGaussians4D  & 0.61 & 0.54 & 0.931 & 0.908 & 0.945 \\
        
        \textbf{Ours} & \textbf{0.74} & \textbf{0.66} & \textbf{0.950} & \textbf{0.948} & \textbf{0.963} \\
        \midrule
        
        Ground Truth & 0.87 & - & - & - & 1.0 \\
        \bottomrule
    \end{tabular}
    }
\end{table}

% \begin{table}[ht] % Use table* and [t] for top placement
%     \centering
%     \caption{\orlF{\textbf{Quantitative Comparison with Baselines.} Our method achieves significantly higher Motion Fidelity and CLIP similarity scores across both benchmarks compared to the baselines. We encourage the reader to refer to Figure~\ref{fig:qualitative_comparison} for a visual illustration of these improvements.}}
%     \label{tab:main_experiment}
%     \resizebox{\columnwidth}{!}{
%     \begin{tabular}{lccc}
%         \toprule
%         & \multicolumn{2}{c}{\textbf{Motion Fidelity} $\uparrow$} & \textbf{CLIP Score} $\uparrow$ \\
%         \cmidrule(lr){2-3} \cmidrule(lr){4-4}
%         & Mini-Mixamo & Cross-Category & Mini-Mixamo \\
%         \midrule
%         SC4D Motion Transfer & 0.6539 & 0.5585 & 0.9045 \\
%         DreamGaussians4D  & 0.6115 & 0.5428 & 0.9451 \\
        
%         \textbf{Ours} & \textbf{0.7384} & \textbf{0.6617} & \textbf{0.9628}  \\
%         \midrule
        
%         GT full dataset & 0.874 & - & 1.0 \\
%         \bottomrule
%     \end{tabular}
%     }
% \end{table}

A quantitative comparison with the adapted baselines is presented in Table~\ref{tab:main_experiment}. Our method outperforms both SC4D and DG4D in Motion Fidelity and CLIP score, demonstrating superior alignment between source and target motions while preserving the target object's structure and identity. To further illustrate these improvements, Figure~\ref{fig:qualitative_comparison} provides a qualitative evaluation, showing (Left) rendered views of the target object and source videos, and (Right) views of the reconstructed dynamic 3DGS outputs from our method and the baselines. As shown, while SC4D captures the correct motion to some extent, its text-based appearance transfer introduces a significant visual discrepancy from the original object. Similarly, DG4D, despite leveraging the target’s original 3DGS, struggles to maintain motion coherence over time, as reflected in both quantitative scores and observable distortions. In contrast, our approach produces visually coherent reconstructions while faithfully transferring the semantic essence of the motion. For example, our method accurately captures the stalking crouch of the skeleton transferred from the panther and the sliding motion of the robotic arm, transferred from a simple cube—demonstrating its ability to generalize across diverse object categories and motion types.
\subsection{In-the-wild Motion Transfer}
\begin{figure}[h]
    \centering
    \includegraphics[width=1.\linewidth]{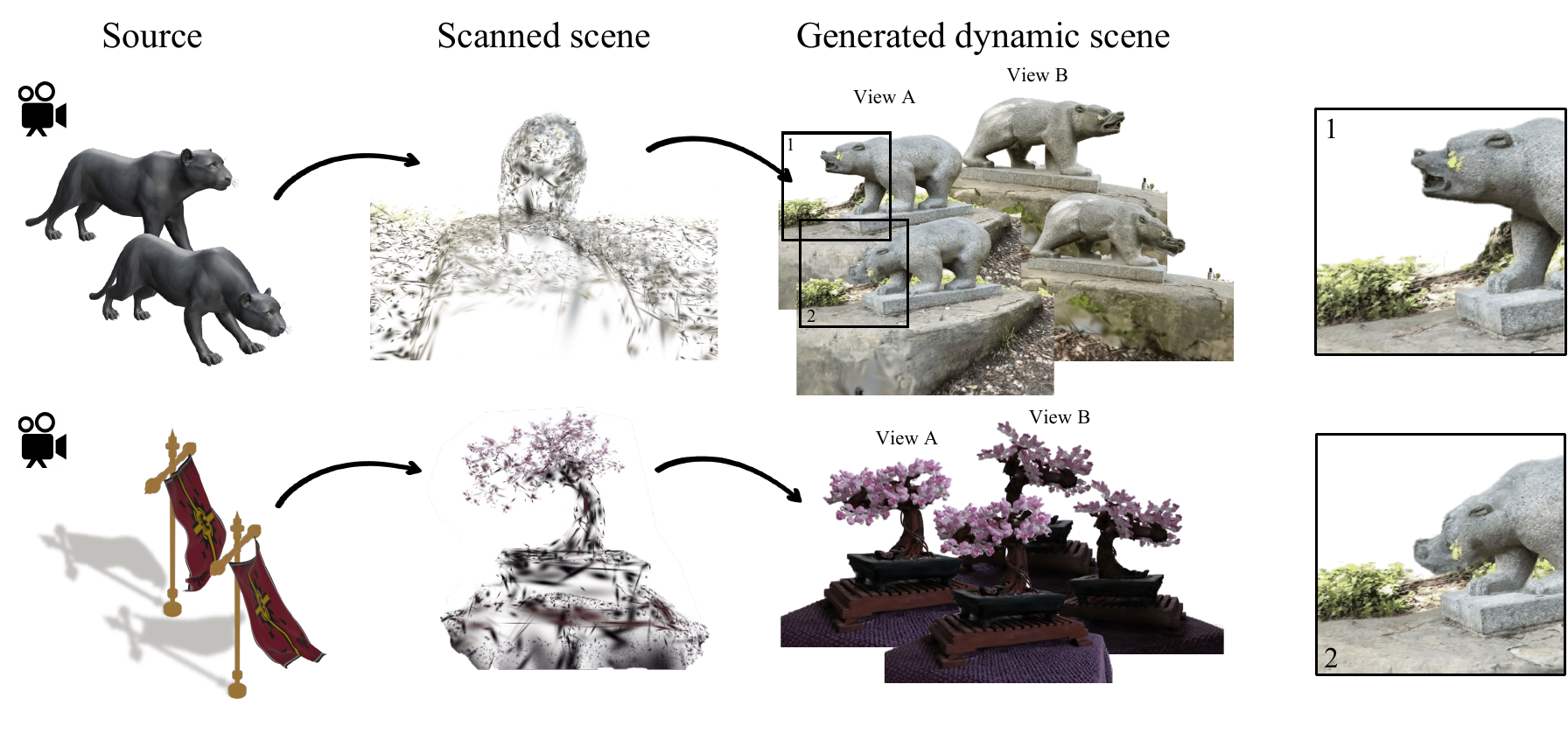}
    \caption{\textbf{In-the-wild Motion Transfer.} Our method animates 3D assets reconstructed from real-world imagery, demonstrating robust semantic motion transfer in real scenes using 3DGS. The ‘Scanned scene’ column visualizes sparse sample of the 3DGS scene produced by our reconstruction stage, highlighting that the motion is applied directly to the scene.}
    \label{fig:inthewild}
\end{figure}

Leveraging a video diffusion prior trained on real-world footage, our model generalizes beyond synthetic settings by enabling the application of motion embeddings to arbitrary 3D targets in real scenes (see Fig.\ref{fig:inthewild}). This capability is increasingly important given the rapid progress in 3DGS, which supports high-quality reconstructions from in-the-wild imagery. As 3DGS pipelines become more accessible and robust for real-world data, the ability to animate such assets with semantically meaningful motion becomes a key step toward practical deployment.

\subsection{Human Preference Study}
\begin{figure}[h]    
    \centering
    \includegraphics[width=1.\linewidth]{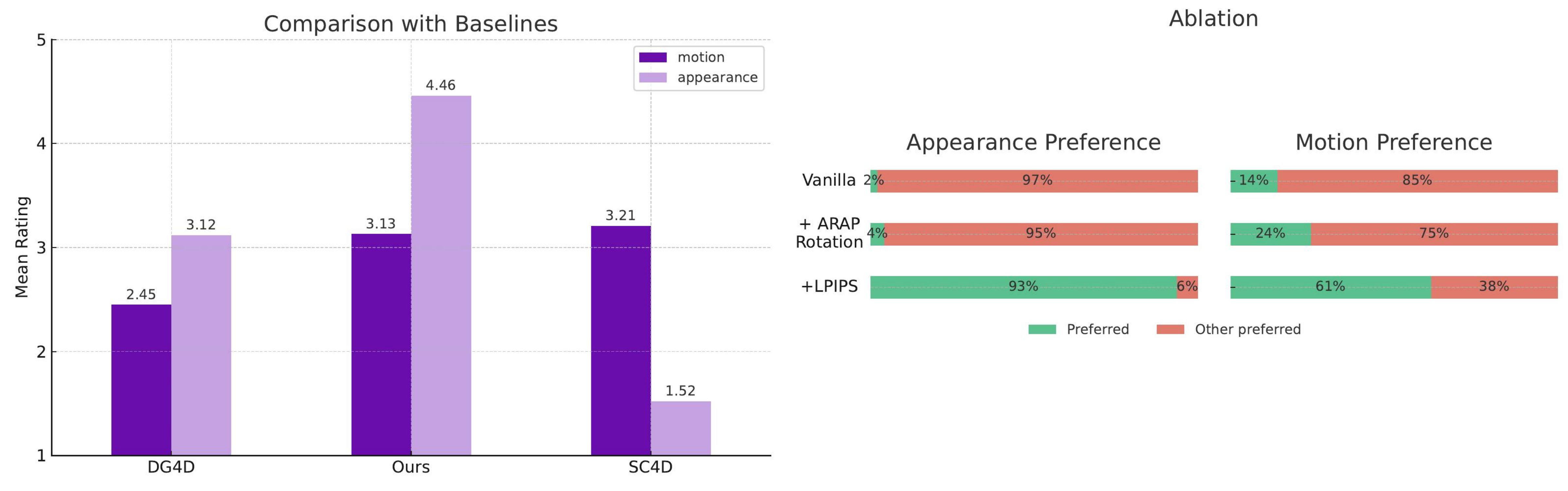}
    \caption{\textbf{Human preference study.} Left: Mean subjective ratings for motion plausibility and appearance fidelity. Our method is the only one to preserve target identity while delivering high-quality motion transfer. Right: Preference results from an ablation study, showing that both LPIPS and ARAP rotation substantially improve perceptual quality.}
    \label{fig:human_survey}
\end{figure}

\noindent We conducted a human preference study to evaluate the perceived quality of semantic motion transfer. Participants compared side-by-side outputs from different methods and rated them based on visual fidelity (i.e., adherence to the 3D asset) and motion plausibility. Results are shown in Fig.~\ref{fig:human_survey}. On the left, our method outperforms adapted baselines in appearance quality, with an average rating of 4.66/5, and is the only one to consistently preserve object identity. For motion quality, our approach matches SC4D—though SC4D fails to maintain appearance—and surpasses DG4D.

On the right, we show an ablation study on regularization. Both ARAP rotation and LPIPS loss improve perceived motion realism and appearance fidelity. Survey examples and interface screenshots are provided in Sec.~\ref{app:human_study} of the supplementary.

\subsection{Novel View Motion Synthesis}
\label{sec:novel_view}
Our anchor-based mechanism is primarily designed to accelerate motion transfer and collaborate motion cues across input views. However, due to its interpolatory nature, it is natural to examine whether it exhibits the emergent property of synthesizing motion embeddings at unseen views. To evaluate this, we interpolate the anchors to create novel-view embeddings and pair them with a single frame from a previously unseen viewpoint to synthesize a novel-view video of the source object. Table ~\ref{tab:novel_view} presents the results. Remarkably, the output exhibits excellent motion consistency and structural integrity. Furthermore, we show that simple interpolation of simple motion embeddings or a single global embedding both fail at this task, underscoring the effectiveness of our approach (See visual comparison in Fig.~\ref{fig:Novel_View}). This capability extends to the target object as well (see Appendix~\ref{app:novel_motion_synthesis}). We hypothesize that this ability could be particularly valuable for viewpoint densification applications. Incorporating this capability into a motion transfer pipeline is an exciting avenue for future work. 

\begin{table}[h]
    \centering
    \caption{Quantitative results for Novel-view Motion Synthesis using MSE and LPIPS. Metrics are computed only on viewing angles unseen during training.}    
    \label{tab:novel_view}
    \resizebox{0.65\linewidth}{!}{
    \begin{tabular}{lcc}
        \toprule
        \textbf{Method} & \textbf{MSE} $\downarrow$ & \textbf{LPIPS} $\downarrow$ \\
        \midrule
        Simple & 0.0111 $\pm$ 0.0034 & 0.1058 $\pm$ 0.0179 \\
        Global & 0.0069 $\pm$ 0.0049 & 0.0649 $\pm$ 0.0306 \\
        Ours & \textbf{0.0028 $\pm$ 0.0016} & \textbf{0.0403 $\pm$ 0.0170} \\
        \bottomrule
    \end{tabular}
    }

\end{table}

% reduce space between figs
\setlength{\textfloatsep}{2pt}
\setlength{\belowcaptionskip}{1pt}
\setlength{\abovecaptionskip}{1pt}

\begin{figure*}[h]
    \centering
    \includegraphics[width=\linewidth]{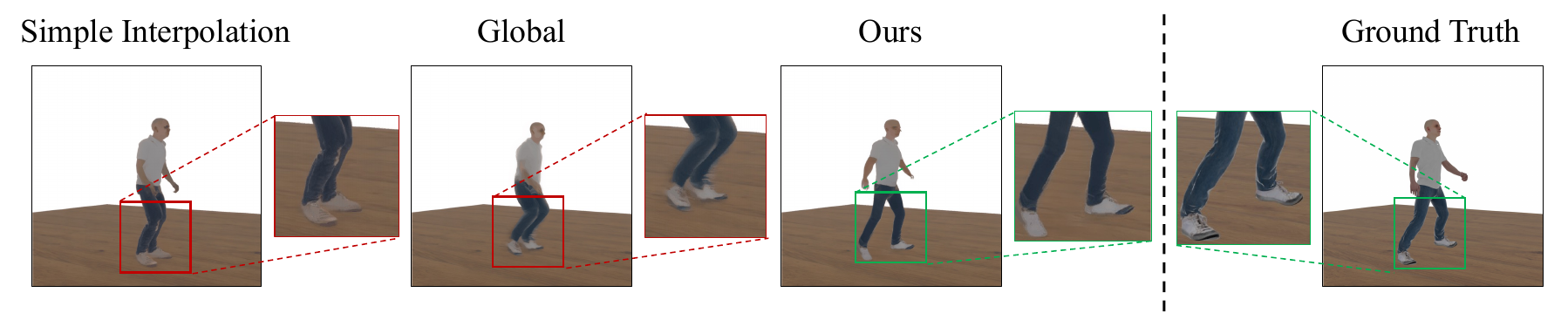}
    \caption{\textbf{Qualitative comparison of novel-view motion synthesis}. 
    Interpolating simple motion embeddings (Eq.~\ref{eq:simple_embeddings}) and a single global embedding both fail to generalize motion embeddings to views unseen during optimization. In contrast, our anchor-based mechanism successfully recovers faithful motion. A similar behavior is observed for the target object before and after reconstruction (Appendix~\ref{app:novel_motion_synthesis}).}
    \label{fig:Novel_View}
\end{figure*}

    % Far-left was generated using interpolation of two simple motion embeddings~\ref{eq:simple_embeddings}, while the Global video was produced using a single global motion embedding. Ours video was generated using the anchor-based mechanism, and the Ground Truth (GT) video is shown on the far right. The presented viewpoint was not seen during training, demonstrating each method’s ability to generalize to unseen views. The video is frozen at frame 4 for direct visual comparison.}
\begin{figure*}[h]
    \centering
    \includegraphics[width=\linewidth]{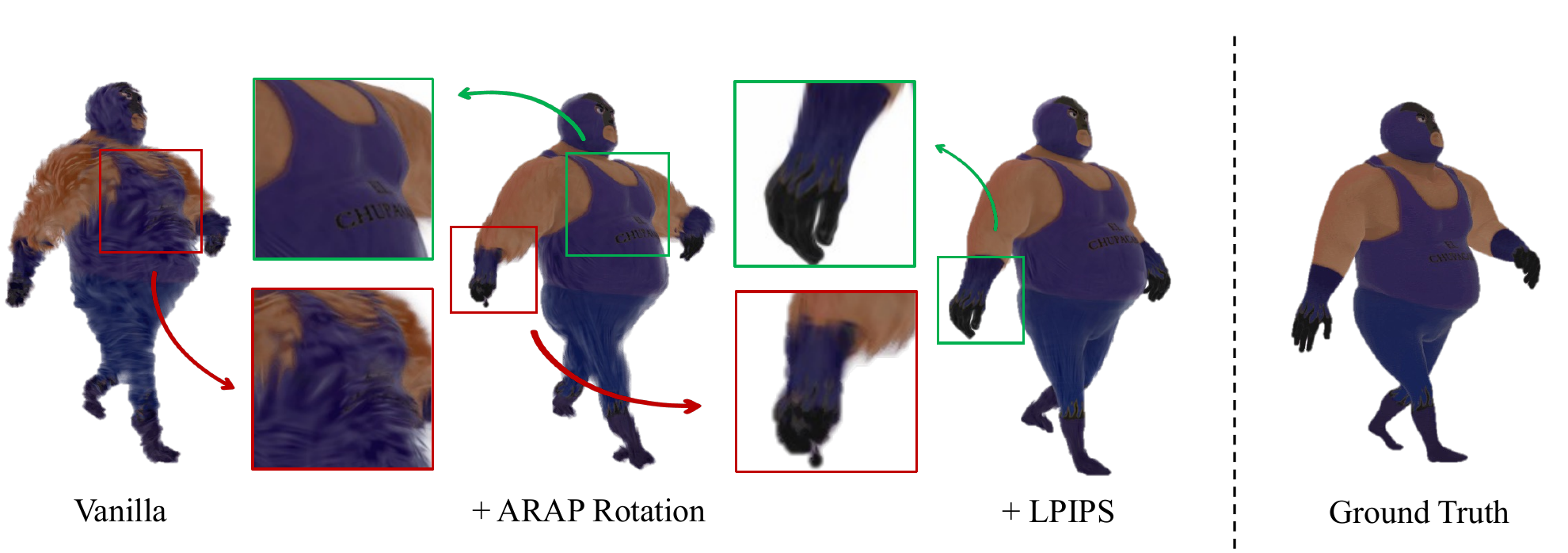}
    \caption{\textbf{Qualitative comparison of reconstruction improvements.} The vanilla reconstruction introduces geometric and texture artifacts, while ARAP rotation and LPIPS regularization significantly enhance reconstruction quality, preserving fine details.}
    \label{fig:ablation_quality}
\end{figure*}

% The vanilla method (left) introduces surface and texture artifacts (red boxes). Applying ARAP Rotation helps retain the original texture, making details such as the text on the shirt more legible (green boxes). Adding LPIPS loss further enhances detail retention, reducing fine details distortions in high-magnitude deformation areas. The Ground Truth serves as a reference for ideal reconstruction
\begin{figure}[h]
    \centering
    \includegraphics[width=1.\linewidth]{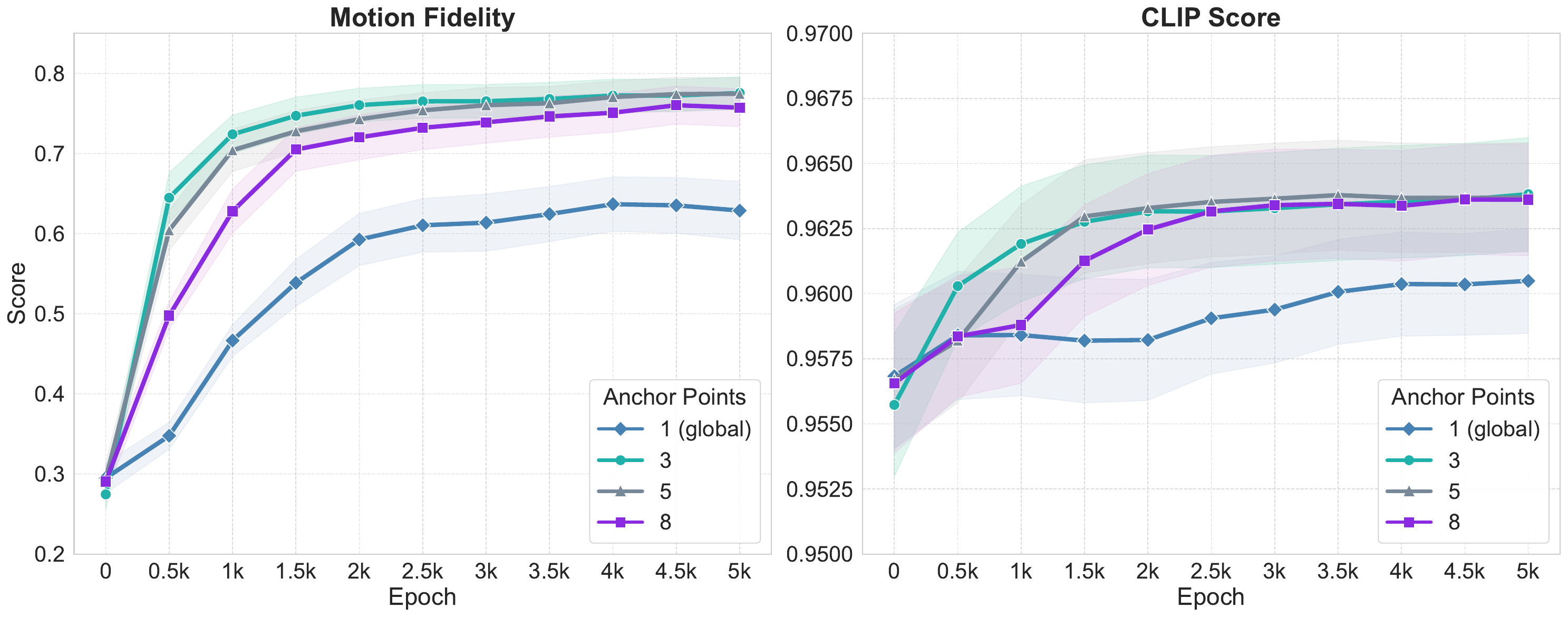}
    \caption{\textbf{Effect of number of anchors.} Reducing anchors speeds up convergence, but too few results in both poor quality and slower optimization. See Section~\ref{sec:ablation} for details.}
    % \orl{what's the main message here? let's see what happens with a single anchor. }  
    \label{fig:ablation_quant}
\end{figure}

% Each point represents the average accuracy over five random seeds, with shaded areas indicating the 95\% confidence intervals. Different colors and markers denote the number of learnable anchors used in training and reconstruction

\setlength{\textfloatsep}{10pt}
\setlength{\belowcaptionskip}{10pt}
\setlength{\abovecaptionskip}{10pt}

\subsection{Ablation Studies}
\label{sec:ablation}

\paragraph{Number of anchors.}
\noindent A key parameter in our method is the number of anchors, \( K \), which trades off convergence speed and reconstruction quality. As shown in Figure~\ref{fig:ablation_quant}, fewer anchors speed up convergence by encouraging broader view consistency, but too few—e.g., a single global embedding—degrade performance and cause motion hallucinations. We choose \( K = 5 \) as a balanced setting. Visual examples and further analysis appear in Appendix~\ref{app:inversion-ablation}.

\paragraph{Regularization terms.} We refine the reconstruction pipeline using ARAP Rotation and LPIPS regularization, which significantly improve structural integrity and fine-detail preservation (Figure~\ref{fig:ablation_quality}). These terms enhance texture stability, maintaining legibility of details like text and intricate patterns. See Appendix~\ref{app:reconstruction-ablation} for details.

\section{Conclusion \& Limitations}
\label{sec:conclusion}
We introduced Gaussian See, Gaussian Do, pioneering the first approach to lift implicit video diffusion motion transfer to 3D, achieving semantic 3D motion transfer from multiview video to 3DGS. We also established the first benchmark for this task, demonstrating superior performance over adapted baselines. However, we identify several limitations as exciting directions for future work. Runtime remains a challenge, as condition inversion is computationally expensive, and while our anchor-based mechanism accelerates convergence, further speed improvements are necessary. Additionally, the lack of a robust 3D semantic motion metric highlights a need for better evaluation tools. Finally, our method shows promise in synthesizing motion at novel views, and we aim to incorporate this capability into the pipeline to further boost performance.

\paragraph{Release \& Licensing.}
We release the code and benchmark under the \emph{BigCode OpenRAIL-M v0.1} license, which explicitly prohibits impersonation and deepfakes without consent, as specified in clause h.

\paragraph*{Acknowledgments.}
 We thank Manuel Kansy for helpful discussions. Or Litany acknowledges the support of the Israel Science Foundation (grant No. 624/25) and the Azrieli Foundation Early Career Faculty Fellowship.

% Bibliography
\bibliographystyle{ACM-Reference-Format}
\bibliography{main}

@String(CVPR= {IEEE Conf. Comput. Vis. Pattern Recog.})

@String(ECCV= {Eur. Conf. Comput. Vis.})

@String(TOG= {ACM Trans. Graph.})

@String(CVPR  = {CVPR})

@String(ECCV  = {ECCV})

@String(TOG   = {ACM TOG})

@inproceedings{aist-dance-db,
           author = {Shuhei Tsuchida and Satoru Fukayama and Masahiro Hamasaki and Masataka Goto}, 
           title = {AIST Dance Video Database: Multi-genre, Multi-dancer, and Multi-camera Database for Dance Information Processing}, 
           booktitle = {Proceedings of the 20th International Society for Music Information Retrieval Conference, {ISMIR} 2019},
           address = {Delft, Netherlands}, 
           year = 2019, 
           month = nov }

@inproceedings{villegas2018neural,
  title={Neural kinematic networks for unsupervised motion retargetting},
  author={Villegas, Ruben and Yang, Jimei and Ceylan, Duygu and Lee, Honglak},
  booktitle={Proceedings of the IEEE conference on computer vision and pattern recognition},
  pages={8639--8648},
  year={2018}
}

@inproceedings{chen2023weakly,
  title={Weakly-supervised 3d pose transfer with keypoints},
  author={Chen, Jinnan and Li, Chen and Lee, Gim Hee},
  booktitle={Proceedings of the IEEE/CVF International Conference on Computer Vision},
  pages={15156--15165},
  year={2023}
}

@article{sun2022human,
  title={Human motion transfer with 3d constraints and detail enhancement},
  author={Sun, Yang-Tian and Fu, Qian-Cheng and Jiang, Yue-Ren and Liu, Zitao and Lai, Yu-Kun and Fu, Hongbo and Gao, Lin},
  journal={IEEE Transactions on Pattern Analysis and Machine Intelligence},
  volume={45},
  number={4},
  pages={4682--4693},
  year={2022},
  publisher={IEEE}
}

@inproceedings{petrovich2022temos,
  title={Temos: Generating diverse human motions from textual descriptions},
  author={Petrovich, Mathis and Black, Michael J and Varol, G{\"u}l},
  booktitle={European Conference on Computer Vision},
  pages={480--497},
  year={2022},
  organization={Springer}
}

@article{goh2021multimodal,
  title={Multimodal neurons in artificial neural networks},
  author={Goh, Gabriel and Cammarata, Nick and Voss, Chelsea and Carter, Shan and Petrov, Michael and Schubert, Ludwig and Radford, Alec and Olah, Chris},
  journal={Distill},
  volume={6},
  number={3},
  pages={e30},
  year={2021}
}

@inproceedings{jain2021putting,
  title={Putting nerf on a diet: Semantically consistent few-shot view synthesis},
  author={Jain, Ajay and Tancik, Matthew and Abbeel, Pieter},
  booktitle={Proceedings of the IEEE/CVF International Conference on Computer Vision},
  pages={5885--5894},
  year={2021}
}

@inproceedings{zheng2024unified,
  title={A unified approach for text-and image-guided 4d scene generation},
  author={Zheng, Yufeng and Li, Xueting and Nagano, Koki and Liu, Sifei and Hilliges, Otmar and De Mello, Shalini},
  booktitle={Proceedings of the IEEE/CVF Conference on Computer Vision and Pattern Recognition},
  pages={7300--7309},
  year={2024}
}

@article{zhao2023animate124,
  title={Animate124: Animating one image to 4d dynamic scene},
  author={Zhao, Yuyang and Yan, Zhiwen and Xie, Enze and Hong, Lanqing and Li, Zhenguo and Lee, Gim Hee},
  journal={arXiv preprint arXiv:2311.14603},
  year={2023}
}

@inproceedings{wang2024animatabledreamer,
  title={Animatabledreamer: Text-guided non-rigid 3d model generation and reconstruction with canonical score distillation},
  author={Wang, Xinzhou and Wang, Yikai and Ye, Junliang and Sun, Fuchun and Wang, Zhengyi and Wang, Ling and Liu, Pengkun and Sun, Kai and Wang, Xintong and Xie, Wende and others},
  booktitle={European Conference on Computer Vision},
  pages={321--339},
  year={2024},
  organization={Springer}
}

@inproceedings{jacobianrigfree,
    author = {Muralikrishnan, Sanjeev and Dutt, Niladri and Chaudhuri, Siddhartha and Aigerman, Noam and Kim, Vladimir and Fisher, Matthew and Mitra, Niloy J.},
    title = {Temporal Residual Jacobians for Rig-Free Motion Transfer},
    year = {2024},
    isbn = {978-3-031-73635-3},
    publisher = {Springer-Verlag},
    address = {Berlin, Heidelberg},
    url = {https://doi.org/10.1007/978-3-031-73636-0_6},
    doi = {10.1007/978-3-031-73636-0_6},
    booktitle = {Computer Vision – ECCV 2024: 18th European Conference, Milan, Italy, September 29–October 4, 2024, Proceedings, Part LVIII},
    pages = {93–109},
    numpages = {17},
    location = {Milan, Italy}
    }

@article{singer2023text4d,
  author = {Singer, Uriel and Sheynin, Shelly and Polyak, Adam and Ashual, Oron and
           Makarov, Iurii and Kokkinos, Filippos and Goyal, Naman and Vedaldi, Andrea and
           Parikh, Devi and Johnson, Justin and Taigman, Yaniv},
  title = {Text-To-4D Dynamic Scene Generation},
  journal = {arXiv:2301.11280},
  year = {2023},
}

@misc{Mixamo,
  author       = {Adobe Systems Inc.},
  title        = {Mixamo},
  year         = 2024,
  url          = {https://doi.org/10.57702/xbooguyb},
  note         = {DOI retrieved: December 2, 2024}
}

@article{zhang2024motiondiffuse,
  title={Motiondiffuse: Text-driven human motion generation with diffusion model},
  author={Zhang, Mingyuan and Cai, Zhongang and Pan, Liang and Hong, Fangzhou and Guo, Xinying and Yang, Lei and Liu, Ziwei},
  journal={IEEE transactions on pattern analysis and machine intelligence},
  volume={46},
  number={6},
  pages={4115--4128},
  year={2024},
  publisher={IEEE}
}

@inproceedings{petrovich2024multi,
  title={Multi-track timeline control for text-driven 3D human motion generation},
  author={Petrovich, Mathis and Litany, Or and Iqbal, Umar and Black, Michael J and Varol, Gul and Bin Peng, Xue and Rempe, Davis},
  booktitle={Proceedings of the IEEE/CVF Conference on Computer Vision and Pattern Recognition},
  pages={1911--1921},
  year={2024}
}

@article{tevet2022human,
  title={Human motion diffusion model},
  author={Tevet, Guy and Raab, Sigal and Gordon, Brian and Shafir, Yonatan and Cohen-Or, Daniel and Bermano, Amit H},
  journal={arXiv preprint arXiv:2209.14916},
  year={2022}
}

@article{chu2024humanrig,
  title={HumanRig: Learning Automatic Rigging for Humanoid Character in a Large Scale Dataset},
  author={Chu, Zedong and Xiong, Feng and Liu, Meiduo and Zhang, Jinzhi and Shao, Mingqi and Sun, Zhaoxu and Wang, Di and Xu, Mu},
  journal={arXiv preprint arXiv:2412.02317},
  year={2024}
}

@article{guo2024make,
  title={Make-It-Animatable: An Efficient Framework for Authoring Animation-Ready 3D Characters},
  author={Guo, Zhiyang and Xiang, Jinxu and Ma, Kai and Zhou, Wengang and Li, Houqiang and Zhang, Ran},
  journal={arXiv preprint arXiv:2411.18197},
  year={2024}
}

@article{liu2025riganything,
  title={RigAnything: Template-Free Autoregressive Rigging for Diverse 3D Assets},
  author={Liu, Isabella and Xu, Zhan and Yifan, Wang and Tan, Hao and Xu, Zexiang and Wang, Xiaolong and Su, Hao and Shi, Zifan},
  journal={arXiv preprint arXiv:2502.09615},
  year={2025}
}

@article{xu2020rignet,
  title={Rignet: Neural rigging for articulated characters},
  author={Xu, Zhan and Zhou, Yang and Kalogerakis, Evangelos and Landreth, Chris and Singh, Karan},
  journal={arXiv preprint arXiv:2005.00559},
  year={2020}
}

@article{li2021learning,
  title={Learning skeletal articulations with neural blend shapes},
  author={Li, Peizhuo and Aberman, Kfir and Hanocka, Rana and Liu, Libin and Sorkine-Hornung, Olga and Chen, Baoquan},
  journal={ACM Transactions on Graphics (TOG)},
  volume={40},
  number={4},
  pages={1--15},
  year={2021},
  publisher={ACM New York, NY, USA}
}

@article{baran2007automatic,
  title={Automatic rigging and animation of 3d characters},
  author={Baran, Ilya and Popovi{\'c}, Jovan},
  journal={ACM Transactions on graphics (TOG)},
  volume={26},
  number={3},
  pages={72--es},
  year={2007},
  publisher={ACM New York, NY, USA}
}

@article{ma2023tarig,
  title={TARig: Adaptive template-aware neural rigging for humanoid characters},
  author={Ma, Jing and Zhang, Dongliang},
  journal={Computers \& Graphics},
  volume={114},
  pages={158--167},
  year={2023},
  publisher={Elsevier}
}

@inproceedings{huang2024sc,
  title={Sc-gs: Sparse-controlled gaussian splatting for editable dynamic scenes},
  author={Huang, Yi-Hua and Sun, Yang-Tian and Yang, Ziyi and Lyu, Xiaoyang and Cao, Yan-Pei and Qi, Xiaojuan},
  booktitle={Proceedings of the IEEE/CVF Conference on Computer Vision and Pattern Recognition},
  pages={4220--4230},
  year={2024}
}

@article{rahamim2024bringing,
  title={Bringing Objects to Life: 4D generation from 3D objects},
  author={Rahamim, Ohad and Malca, Ori and Samuel, Dvir and Chechik, Gal},
  journal={arXiv preprint arXiv:2412.20422},
  year={2024}
}

@article{gal2022image,
  title={An image is worth one word: Personalizing text-to-image generation using textual inversion},
  author={Gal, Rinon and Alaluf, Yuval and Atzmon, Yuval and Patashnik, Or and Bermano, Amit H and Chechik, Gal and Cohen-Or, Daniel},
  journal={arXiv preprint arXiv:2208.01618},
  year={2022}
}

@article{kerbl20233d,
  title={3d gaussian splatting for real-time radiance field rendering.},
  author={Kerbl, Bernhard and Kopanas, Georgios and Leimk{\"u}hler, Thomas and Drettakis, George},
  journal={ACM Trans. Graph.},
  volume={42},
  number={4},
  pages={139--1},
  year={2023}
}

@inproceedings{yatim2024space,
  title={Space-time diffusion features for zero-shot text-driven motion transfer},
  author={Yatim, Danah and Fridman, Rafail and Bar-Tal, Omer and Kasten, Yoni and Dekel, Tali},
  booktitle={Proceedings of the IEEE/CVF Conference on Computer Vision and Pattern Recognition},
  pages={8466--8476},
  year={2024}
}

@article{raab2023single,
  title={Single motion diffusion},
  author={Raab, Sigal and Leibovitch, Inbal and Tevet, Guy and Arar, Moab and Bermano, Amit H and Cohen-Or, Daniel},
  journal={arXiv preprint arXiv:2302.05905},
  year={2023}
}

@inproceedings{da2022dual,
  title={Dual-head contrastive domain adaptation for video action recognition},
  author={Da Costa, Victor G Turrisi and Zara, Giacomo and Rota, Paolo and Oliveira-Santos, Thiago and Sebe, Nicu and Murino, Vittorio and Ricci, Elisa},
  booktitle={Proceedings of the IEEE/CVF Winter Conference on Applications of Computer Vision},
  pages={1181--1190},
  year={2022}
}

@inproceedings{wang2025towards,
  title={Towards High-Quality 3D Motion Transfer with Realistic Apparel Animation},
  author={Wang, Rong and Mao, Wei and Lu, Changsheng and Li, Hongdong},
  booktitle={European Conference on Computer Vision},
  pages={35--51},
  year={2025},
  organization={Springer}
}

@inproceedings{raab2024monkey,
  title={Monkey see, monkey do: Harnessing self-attention in motion diffusion for zero-shot motion transfer},
  author={Raab, Sigal and Gat, Inbar and Sala, Nathan and Tevet, Guy and Shalev-Arkushin, Rotem and Fried, Ohad and Bermano, Amit Haim and Cohen-Or, Daniel},
  booktitle={SIGGRAPH Asia 2024 Conference Papers},
  pages={1--13},
  year={2024}
}

@inproceedings{wu2024reconfusion,
  title={Reconfusion: 3d reconstruction with diffusion priors},
  author={Wu, Rundi and Mildenhall, Ben and Henzler, Philipp and Park, Keunhong and Gao, Ruiqi and Watson, Daniel and Srinivasan, Pratul P and Verbin, Dor and Barron, Jonathan T and Poole, Ben and others},
  booktitle={Proceedings of the IEEE/CVF conference on computer vision and pattern recognition},
  pages={21551--21561},
  year={2024}
}

@article{li2024dreammesh4d,
  title={Dreammesh4d: Video-to-4d generation with sparse-controlled gaussian-mesh hybrid representation},
  author={Li, Zhiqi and Chen, Yiming and Liu, Peidong},
  journal={Advances in Neural Information Processing Systems},
  volume={37},
  pages={21377--21400},
  year={2024}
}

@inproceedings{kim2024most,
  title={MoST: Motion Style Transformer between Diverse Action Contents},
  author={Kim, Boeun and Kim, Jungho and Chang, Hyung Jin and Choi, Jin Young},
  booktitle={Proceedings of the IEEE/CVF Conference on Computer Vision and Pattern Recognition},
  pages={1705--1714},
  year={2024}
}

@article{wu2024sc4d,
  author = {Wu, Zijie and Yu, Chaohui and Jiang, Yanqin and Cao, Chenjie and Wang Fan and Bai, Xiang.},
  title = {SC4D: Sparse-Controlled Video-to-4D Generation and Motion Transfer},
  journal = {arxiv:2404.03736},
  year = {2024},
}

@article{kansy2024reenact,
  title={Reenact Anything: Semantic Video Motion Transfer Using Motion-Textual Inversion},
  author={Kansy, Manuel and Naruniec, Jacek and Schroers, Christopher and Gross, Markus and Weber, Romann M},
  journal={arXiv preprint arXiv:2408.00458},
  year={2024}
}

@article{xiao2024video,
  title={Video Diffusion Models are Training-free Motion Interpreter and Controller},
  author={Xiao, Zeqi and Zhou, Yifan and Yang, Shuai and Pan, Xingang},
  journal={arXiv preprint arXiv:2405.14864},
  year={2024}
}

@article{ling2024motionclone,
  title={MotionClone: Training-Free Motion Cloning for Controllable Video Generation},
  author={Ling, Pengyang and Bu, Jiazi and Zhang, Pan and Dong, Xiaoyi and Zang, Yuhang and Wu, Tong and Chen, Huaian and Wang, Jiaqi and Jin, Yi},
  journal={arXiv preprint arXiv:2406.05338},
  year={2024}
}

@misc{wang2024motioninversionvideocustomization,
          title={Motion Inversion for Video Customization}, 
          author={Luozhou Wang and Ziyang Mai and Guibao Shen and Yixun Liang and Xin Tao and Pengfei Wan and Di Zhang and Yijun Li and Yingcong Chen},
          year={2024},
          eprint={2403.20193},
          archivePrefix={arXiv},
          primaryClass={cs.CV},
          url={https://arxiv.org/abs/2403.20193}, 
}

@inproceedings{wei2024dreamvideo,
	title={DreamVideo: Composing Your Dream Videos with Customized Subject and Motion},
	author={Wei, Yujie and Zhang, Shiwei and Qing, Zhiwu and Yuan, Hangjie and Liu, Zhiheng and Liu, Yu and Zhang, Yingya and Zhou, Jingren and Shan, Hongming},
	booktitle={Proceedings of the IEEE/CVF Conference on Computer Vision and Pattern Recognition},
	pages={6537--6549},
	year={2024}
}

@article{zhao2023motiondirector,
          title={MotionDirector: Motion Customization of Text-to-Video Diffusion Models},
          author={Zhao, Rui and Gu, Yuchao and Wu, Jay Zhangjie and Zhang, David Junhao and Liu, Jiawei and Wu, Weijia and Keppo, Jussi and Shou, Mike Zheng},
          journal={arXiv preprint arXiv:2310.08465},
          year={2023}
}

@article{materzynska2023customizing,
              title={Customizing Motion in Text-to-Video Diffusion Models},
              author={Materzy\'nska, Joanna and Sivic, Josef and Shechtman, Eli and Torralba, Antonio and Zhang, Richard and Russell, Bryan},
              journal={arXiv preprint arXiv:2312.04966},
              year={2023}
}

@article{blattmann2023stable,
  title={Stable video diffusion: Scaling latent video diffusion models to large datasets},
  author={Blattmann, Andreas and Dockhorn, Tim and Kulal, Sumith and Mendelevitch, Daniel and Kilian, Maciej and Lorenz, Dominik and Levi, Yam and English, Zion and Voleti, Vikram and Letts, Adam and others},
  journal={arXiv preprint arXiv:2311.15127},
  year={2023}
}

@misc{guo2024animatediffanimatepersonalizedtexttoimage,
      title={AnimateDiff: Animate Your Personalized Text-to-Image Diffusion Models without Specific Tuning}, 
      author={Yuwei Guo and Ceyuan Yang and Anyi Rao and Zhengyang Liang and Yaohui Wang and Yu Qiao and Maneesh Agrawala and Dahua Lin and Bo Dai},
      year={2024},
      eprint={2307.04725},
      archivePrefix={arXiv},
      primaryClass={cs.CV},
      url={https://arxiv.org/abs/2307.04725}, 
}

@article{xing2023dynamicrafter,
  title={DynamiCrafter: Animating Open-domain Images with Video Diffusion Priors},
  author={Xing, Jinbo and Xia, Menghan and Zhang, Yong and Chen, Haoxin and Yu, Wangbo and Liu, Hanyuan and Wang, Xintao and Wong, Tien-Tsin and Shan, Ying},
  journal={arXiv preprint arXiv:2310.12190},
  year={2023}
}

@misc{zhang2023i2vgenxlhighqualityimagetovideosynthesis,
      title={I2VGen-XL: High-Quality Image-to-Video Synthesis via Cascaded Diffusion Models}, 
      author={Shiwei Zhang and Jiayu Wang and Yingya Zhang and Kang Zhao and Hangjie Yuan and Zhiwu Qin and Xiang Wang and Deli Zhao and Jingren Zhou},
      year={2023},
      eprint={2311.04145},
      archivePrefix={arXiv},
      primaryClass={cs.CV},
      url={https://arxiv.org/abs/2311.04145}, 
}

@misc{wang2023modelscopetexttovideotechnicalreport,
      title={ModelScope Text-to-Video Technical Report}, 
      author={Jiuniu Wang and Hangjie Yuan and Dayou Chen and Yingya Zhang and Xiang Wang and Shiwei Zhang},
      year={2023},
      eprint={2308.06571},
      archivePrefix={arXiv},
      primaryClass={cs.CV},
      url={https://arxiv.org/abs/2308.06571}, 
}

@inproceedings{ling2024alignyourgaussians,
    title={Align Your Gaussians: Text-to-4D with Dynamic 3D Gaussians and Composed Diffusion Models},
    author={Ling, Huan and Kim, Seung Wook and Torralba, Antonio and Fidler, Sanja and Kreis, Karsten},
    booktitle={IEEE Conference on Computer Vision and Pattern Recognition ({CVPR})},
    year={2024}
}

@article{bah2024tc4d,
  author = {Bahmani, Sherwin and Liu, Xian and Yifan, Wang and Skorokhodov, Ivan and Rong, Victor and Liu, Ziwei and Liu, Xihui and Park, Jeong Joon and Tulyakov, Sergey and Wetzstein, Gordon and Tagliasacchi, Andrea and Lindell, David B.},
  title = {TC4D: Trajectory-Conditioned Text-to-4D Generation},
  journal = {arXiv},
  year = {2024},
}

@article{bah20244dfy,
  author = {Bahmani, Sherwin and Skorokhodov, Ivan and Rong, Victor and Wetzstein, Gordon and Guibas, Leonidas and Wonka, Peter and Tulyakov, Sergey and Park, Jeong Joon and Tagliasacchi, Andrea and Lindell, David B.},
  title = {4D-fy: Text-to-4D Generation Using Hybrid Score Distillation Sampling},
  journal = {IEEE Conference on Computer Vision and Pattern Recognition ({CVPR})},
  year = {2024},
}

@article{uzolas2024motiondreamer,
  title={MotionDreamer: Zero-Shot 3D Mesh Animation from Video Diffusion Models},
  author={Uzolas, Lukas and Eisemann, Elmar and Kellnhofer, Petr},
  journal={arXiv preprint arXiv:2405.20155},
  year={2024}
}

@inproceedings{
jiang2024consistentd,
title={Consistent4D: Consistent 360{\textdegree} Dynamic Object Generation from Monocular Video},
author={Yanqin Jiang and Li Zhang and Jin Gao and Weiming Hu and Yao Yao},
booktitle={The Twelfth International Conference on Learning Representations},
year={2024},
url={https://openreview.net/forum?id=sPUrdFGepF}
}

@misc{zeng2024stag4dspatialtemporalanchoredgenerative,
      title={STAG4D: Spatial-Temporal Anchored Generative 4D Gaussians}, 
      author={Yifei Zeng and Yanqin Jiang and Siyu Zhu and Yuanxun Lu and Youtian Lin and Hao Zhu and Weiming Hu and Xun Cao and Yao Yao},
      year={2024},
      eprint={2403.14939},
      archivePrefix={arXiv},
      primaryClass={cs.CV},
      url={https://arxiv.org/abs/2403.14939}, 
}

@misc{miao2024pla4d,
      title={PLA4D: Pixel-Level Alignments for Text-to-4D Gaussian Splatting}, 
      author={Qiaowei Miao and Yawei Luo and Yi Yang},
      year={2024},
      eprint={2405.19957},
      archivePrefix={arXiv},
      primaryClass={cs.CV}
}

@inproceedings{luiten2023dynamic,
  title={Dynamic 3D Gaussians: Tracking by Persistent Dynamic View Synthesis},
  author={Luiten, Jonathon and Kopanas, Georgios and Leibe, Bastian and Ramanan, Deva},
  booktitle={3DV},
  year={2024}
}

@inproceedings{sorkine2007rigid,
  title={As-rigid-as-possible surface modeling},
  author={Sorkine, Olga and Alexa, Marc},
  booktitle={Symposium on Geometry processing},
  volume={4},
  pages={109--116},
  year={2007},
  organization={Citeseer}
}

@inproceedings{instructnerf2023,
         author = {Haque, Ayaan and Tancik, Matthew and Efros, Alexei and Holynski, Aleksander and Kanazawa, Angjoo},
         title = {Instruct-NeRF2NeRF: Editing 3D Scenes with Instructions},
         booktitle = {Proceedings of the IEEE/CVF International Conference on Computer Vision},
         year = {2023},
        }

@article{poole2022dreamfusion,
  author = {Poole, Ben and Jain, Ajay and Barron, Jonathan T. and Mildenhall, Ben},
  title = {DreamFusion: Text-to-3D using 2D Diffusion},
  journal = {arXiv},
  year = {2022},
}

@inproceedings{zhang2023skinned,
  title={Skinned Motion Retargeting with Residual Perception of Motion Semantics \& Geometry},
  author={Zhang, Jiaxu and Weng, Junwu and Kang, Di and Zhao, Fang and Huang, Shaoli and Zhe, Xuefei and Bao, Linchao and Shan, Ying and Wang, Jue and Tu, Zhigang},
  booktitle={Proceedings of the IEEE/CVF Conference on Computer Vision and Pattern Recognition},
  pages={13864--13872},
  year={2023}
}

@article{aberman2020skeleton,
  author = {Aberman, Kfir and Li, Peizhuo and Lischinski, Dani and Sorkine-Hornung, Olga and Cohen-Or, Daniel and Chen, Baoquan},
  title = {Skeleton-Aware Networks for Deep Motion Retargeting},
  journal = {ACM Transactions on Graphics (TOG)},
  volume = {39},
  number = {4},
  pages = {62},
  year = {2020},
  publisher = {ACM}
}

@inproceedings{zhang2024generative,
  title={Generative motion stylization of cross-structure characters within canonical motion space},
  author={Zhang, Jiaxu and Chen, Xin and Yu, Gang and Tu, Zhigang},
  booktitle={Proceedings of the 32nd ACM International Conference on Multimedia},
  pages={7018--7026},
  year={2024}
}

@inproceedings{zhang2018perceptual,
  title={The Unreasonable Effectiveness of Deep Features as a Perceptual Metric},
  author={Zhang, Richard and Isola, Phillip and Efros, Alexei A and Shechtman, Eli and Wang, Oliver},
  booktitle={CVPR},
  year={2018}
}

@inproceedings{Radford2021LearningTV,
  title={Learning Transferable Visual Models From Natural Language Supervision},
  author={Alec Radford and Jong Wook Kim and Chris Hallacy and Aditya Ramesh and Gabriel Goh and Sandhini Agarwal and Girish Sastry and Amanda Askell and Pamela Mishkin and Jack Clark and Gretchen Krueger and Ilya Sutskever},
  booktitle={International Conference on Machine Learning},
  year={2021},
  url={https://api.semanticscholar.org/CorpusID:231591445}
}

@article{sumner2007embedded,
author = {Sumner, Robert W. and Schmid, Johannes and Pauly, Mark},
title = {Embedded deformation for shape manipulation},
year = {2007},
issue_date = {July 2007},
publisher = {Association for Computing Machinery},
address = {New York, NY, USA},
volume = {26},
number = {3},
issn = {0730-0301},
url = {https://doi.org/10.1145/1276377.1276478},
doi = {10.1145/1276377.1276478},
journal = {ACM Trans. Graph.},
month = jul,
pages = {80–es},
numpages = {8}
}

@article{ren2023dreamgaussian4d,
  title={DreamGaussian4D: Generative 4D Gaussian Splatting},
  author={Ren, Jiawei and Pan, Liang and Tang, Jiaxiang and Zhang, Chi and Cao, Ang and Zeng, Gang and Liu, Ziwei},
  journal={arXiv preprint arXiv:2312.17142},
  year={2023}
}

% Appendix
\clearpage
\appendix

\section{Additional details on Benchmark}
\label{app:benchmark}
\subsection{Mini-Mixamo Rig-Less benchmark} 
To facilitate the evaluation of 3D rigless motion transfer, we introduce a benchmark derived from the Mixamo dataset \cite{Mixamo}. 
While motion transfer has been extensively studied in rigged settings, no existing benchmark enables controlled and quantitative assessment of motion transfer in a rig-less setup.
Our benchmarks aims to bridge this gap by providing a standardized evaluation protocol for future research in this under-explored domain.
The core challenge in evaluating motion transfer lies in the need for paired data: a source object performing a given motion and a ground-truth target object executing the same motion. 
This pairing allows computation of transfer accuracy by measuring discrepancies between the predicted and actual target motion. Such pairing naturally exists in the Mixamo dataset, which is already a widely used benchmark in the rigged motion transfer domain \cite{zhang2023skinned, da2022dual, raab2023single}.
Leveraging this property, we construct our benchmark by selecting two source figures performing ten distinct motions and ten target figures, for which we provide both their static canonical poses and their ground-truth motion executions.

\subsection{Web-Crawled Inter- and Cross-Category Dataset}

A key aspect of 3D cross-domain motion transfer is enabling motion adaptation across diverse, rigless objects without requiring predefined skeletal structures. Existing datasets rely heavily on Mixamo, which is limited in diversity and requires rigging, restricting the scope of motion transfer benchmarks. In contrast, our dataset removes this constraint, introducing motion transfer for a wider variety of human and non-human targets. This necessitates an unsupervised evaluation approach, for which we use the Motion Fidelity metric to assess motion plausibility as well as target identity preserving. In doing so, we establish a benchmark for this setting.\\
Our dataset serves two main purposes:

\paragraph{Inter-Category Diverse Motion Tranfer.} By eliminating the reliance on rigged characters, we provide a more diverse and challenging evaluation set for human-to-human and animal-to-animal motion transfer. Unlike prior datasets constrained to Mixamo’s limited rigged models, ours enables motion adaptation to a broader range of rigless human representations. 

\paragraph{Cross-Category Motion Transfer.} We extend motion transfer beyond human-to-human interactions, covering human-to-“human-like” and animal-to-“animal-like” motion transfer. While these categories contain fewer target objects, they introduce meaningful diversity beyond conventional datasets.\\
To construct the dataset, we sourced free, publicly available static 3D objects from the web and preprocessed them into a 3D Gaussian Splatting (3DGS) format. This preprocessing step was essential for our setup. Additionally, we built our multi-view motion sequences using free, publicly available animated 3D objects, rendering them into multi-view videos. Our dataset includes 31 diverse motion sequences applied to 33 rigless 3DGS target objects, covering humans, animals, and objects with semantic correspondence to humans or animals. All assets were independently curated to ensure a diverse and challenging dataset.

\subsection{Data Production - Mini-Mixamo}
\label{sec:supp_mixamo_production}
Our dataset is derived from a portion of the Mixamo dataset and follows a structured pipeline to generate both source and ground truth videos, as well as target object reconstructions. Below, we outline the steps taken to construct the dataset.
Two characters, Brian and Megan, were chosen as source figures, while ten target characters were used: Aj, Amy, Castle Guard 01, Crypto, Eve, Knight, Megan, Michelle, Mousey, Ortiz, and Sporty Granny.
 
To ensure consistency, all source and target objects were aligned in a canonical pose with their right foot toe set at . Additionally, a 10x10 plane with a wooden texture (laminate\_floor\_03 from PolyHaven) was added at position (0,0, z canon) beneath the target figures. For rendering, fourteen frames of the source figures were captured along with the wooden plane to create the source videos for each motion. Similarly, fourteen frames of both the source and target figures were rendered without the wooden plane to generate the ground truth videos.

For the reconstruction of target objects, we rendered 100 images per object using the BlenderNeRF plugin in Blender. The camera was set at a radius of 4, and a global exposure of 3.5 was applied to maintain visual consistency. These rendered images were then used to train each object with the original 3DGS repository for 30,000 iterations, with the maximum spherical harmonic order (max\_sh) set to 0.

This structured approach ensured consistency across our dataset, allowing for accurate comparisons and evaluations in our experiments.

\subsection{Data Production - Web Crawled}
Our dataset includes a diverse range of motions and target objects, encompassing human, animal, and inanimate subjects. In total, we incorporated thirteen human motions, with the majority (eleven) sourced from the Mixamo benchmark and the remaining two from Sketchfab. Additionally, we included eleven animal-specific motions, all sourced from Sketchfab, and seven motions corresponding to objects.

The dataset also features a variety of targets across different categories. For animal-related targets, we included twelve instances, ranging from skeleton models to animal toys, all acquired from Sketchfab. Additionally, we incorporated six animal-like object targets, primarily consisting of robotic figures and uniquely shaped tables. For human-related targets, we gathered ten different models from Sketchfab, including humanoid robots and figures demonstrating secondary motion characteristics. Furthermore, we introduced five human-like object targets, such as clothing items, robotic arms, and wooden cross structures.

This broad range of motions and targets ensures diversity in our dataset, allowing for extensive evaluation of semantic motion transfer and adaptation across different entities.

As in \ref{sec:supp_mixamo_production}, we rendered 100 images per object using the BlenderNeRF plugin in Blender. These rendered images were then used to train each object with the original 3DGS repository for 30,000 iterations, with the maximum spherical harmonic order (max sh) set to 0. All of these reconstructed objects were then manually scaled and rotated to ensure pose and scale consistency across the dataset.

\section{Method}
\label{app:method}
\subsection{Implementation details}
Our model uses 5 anchors, 3000 inversion iterations, and motion embeddings of size 15×5×1024. As a backbone, the generative video model is the SVD model (stabilityai/stable-video-diffusion-img2vid). With control points radius of $\beta=0.0745$ and 512 control points initialized via FPS. The MLP structure encodes input time into a 12D vector and control points into a 60D vector using positional encoding. These are concatenated, projected to 256D via a linear layer with ReLU, followed by a 4-layer 256→256 ReLU loop. A residual connection concatenates the input, followed by a 256D linear layer with ReLU, a 2-layer 256→256 ReLU loop, and a final linear layer reducing output to 3D for deformation prediction.

\label{app:implementation}
\subsection{Curriculum Learning }
\label{app:cuririculum}

\begin{figure}
    \centering
    \includegraphics[width=\linewidth]{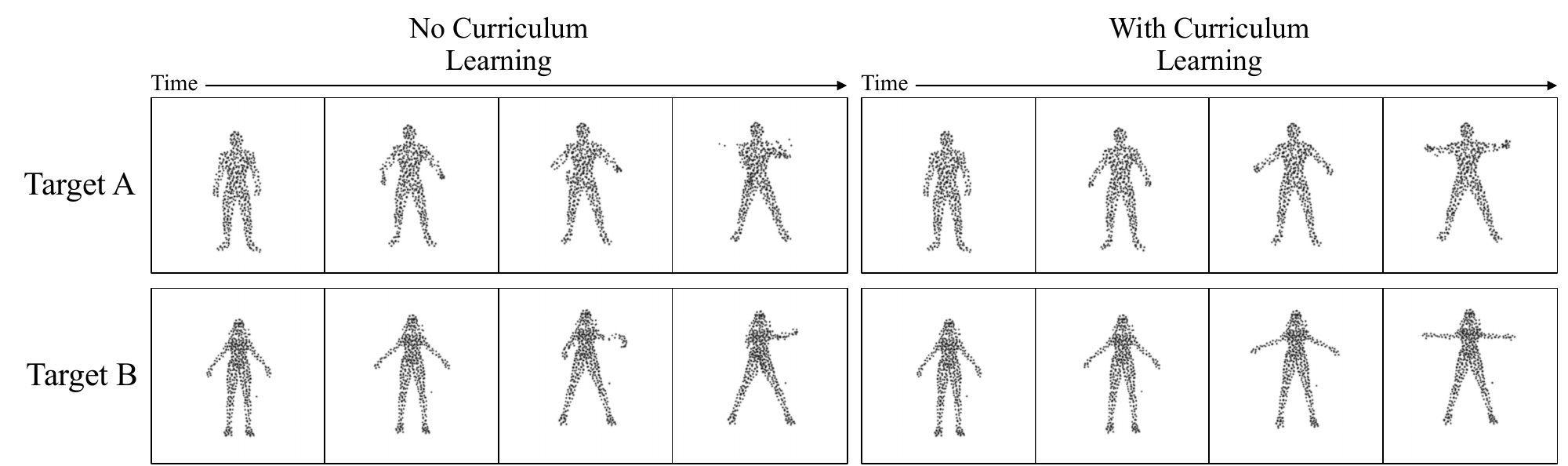}
    \caption{\textbf{Qualitative comparison of motion learning with and without curriculum learning.} The left side demonstrates the control points of a dynamic object after 4D consolidation without curriculum learning, where the model struggles to generalize and produces broken transitions. The right side also shows the control points of a dynamic object, but with curriculum learning durin consolidation, leading to smoother and more natural motion transitions. This effect is particularly significant for cases with large motion variations.}
    \label{fig:curriculum}
\end{figure}

One optional component we experimented with, to attempt and fix "hard" examples is Curriculum Learning.
During training, we noticed that the model struggles to directly learn poses that deviate significantly from the canonical target shape. Furthermore, it tends to overfit to these challenging poses, leading to discrete, step-like motion rather than a smooth, continuous transition. To address this, we introduce curriculum learning, where motion is gradually incorporated over time. Initially, the model learns only small deformations, focusing on local consistency. As training progresses, the allowed deformation range expands, enabling the model to generalize to larger motion variations while maintaining smooth, coherent movement. We note that while curriculum learning plays a crucial role in certain cases, its impact is primarily significant when the magnitude of motion change is large, requiring the model to handle substantial deformations. In cases with smaller motion variations, it is mostly unnecessary. Figure~\ref{fig:curriculum} showcase qualitative analysis of one of the fail cases curriculum fix.

\subsection{Noisy Supervisions}
\label{app:noisy_supervisions}
\begin{figure}
    \centering
    \includegraphics[width=\linewidth]{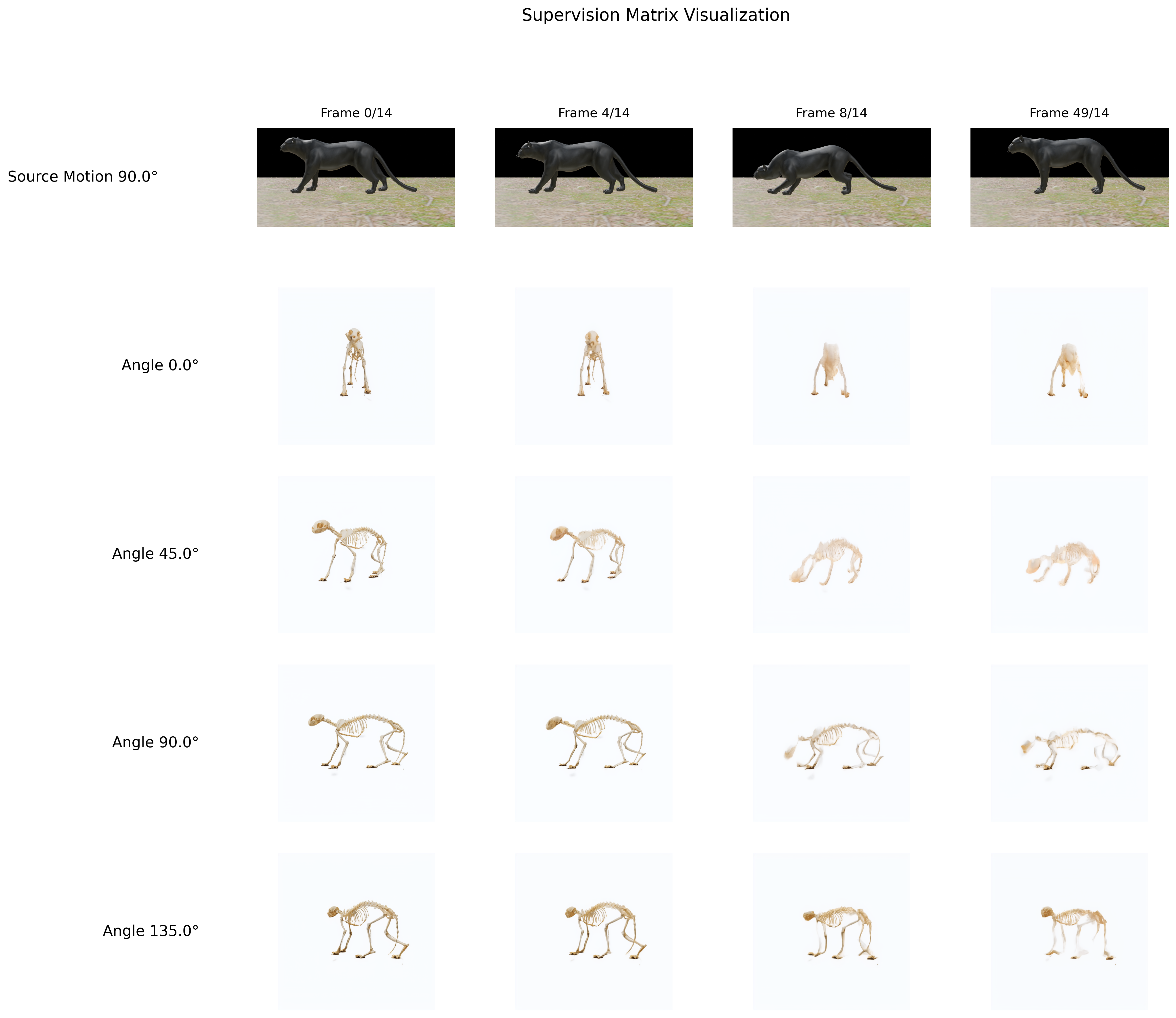}
    \caption{\textbf{Supervision videos example 1.} These are the videos generated by the video model, conditioned on the interpolated motion embedding and on the rendered first frame. They are being used as the supervision videos during the 4D consolidation stage of our method.}\label{fig:noisy_supervision_1}
\end{figure}

\begin{figure}
    \centering
    \includegraphics[width=\linewidth]{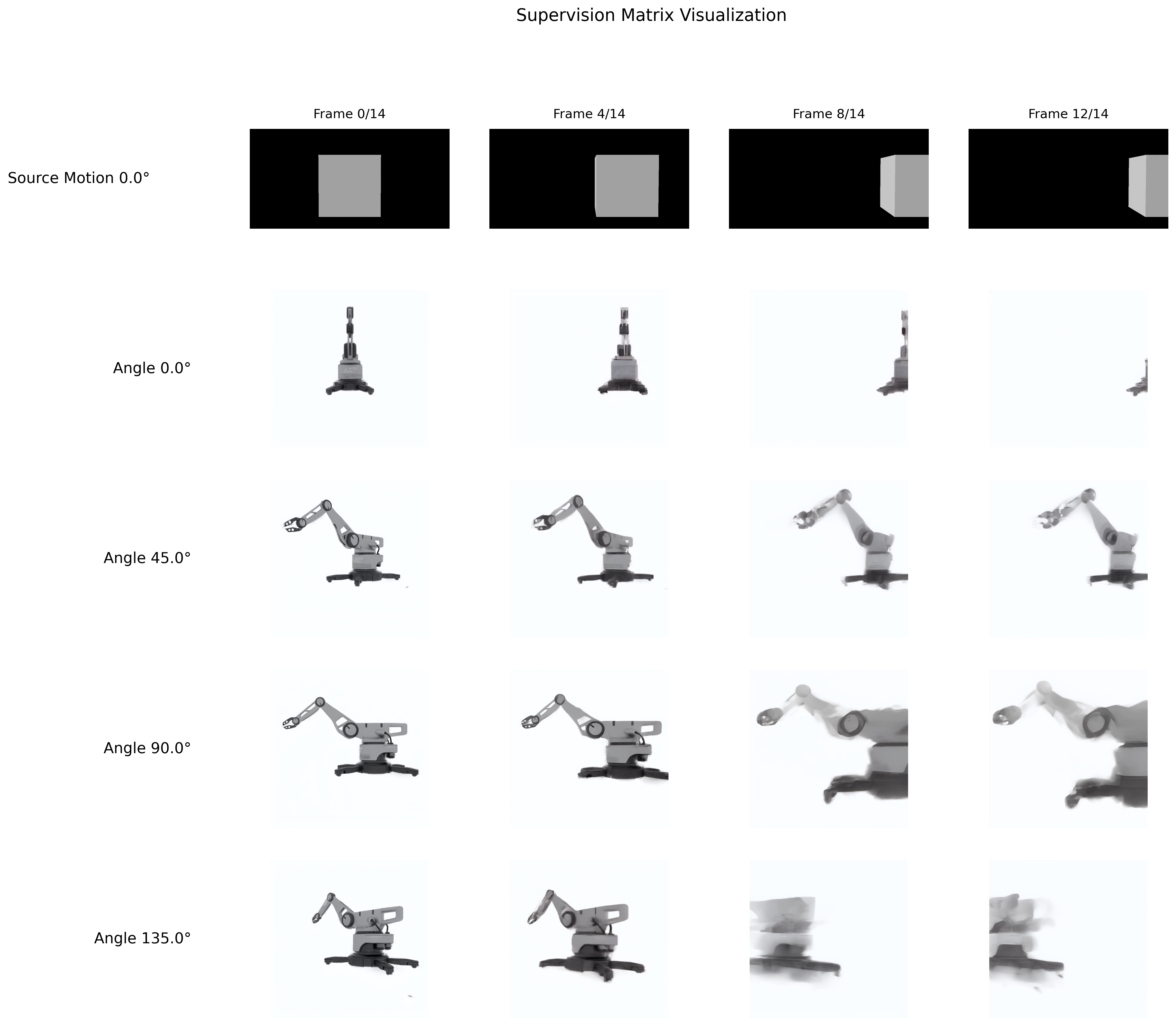}
    \caption{\textbf{Supervision videos example 2.} These are the videos generated by the video model, conditioned on the interpolated motion embedding and on the rendered first frame. They are being used as the supervision videos during the 4D consolidation stage of our method.}\label{fig:noisy_supervision_2}
\end{figure}

To better illustrate the noisy 2D supervisions generated by the Stable Diffusion model, we provide two examples of the resulting videos. Each figure presents a matrix in which each column corresponds to a video frame, progressing from left to right. The first row displays a single-view recording of the source motion, followed by multiple rows showing 2D supervision videos from different viewpoints, generated by applying the motion to the target \ref{fig:noisy_supervision_1},\ref{fig:noisy_supervision_2}.

\subsection{Failure Cases}
\begin{figure}
    \centering
    \includegraphics[width=\linewidth]{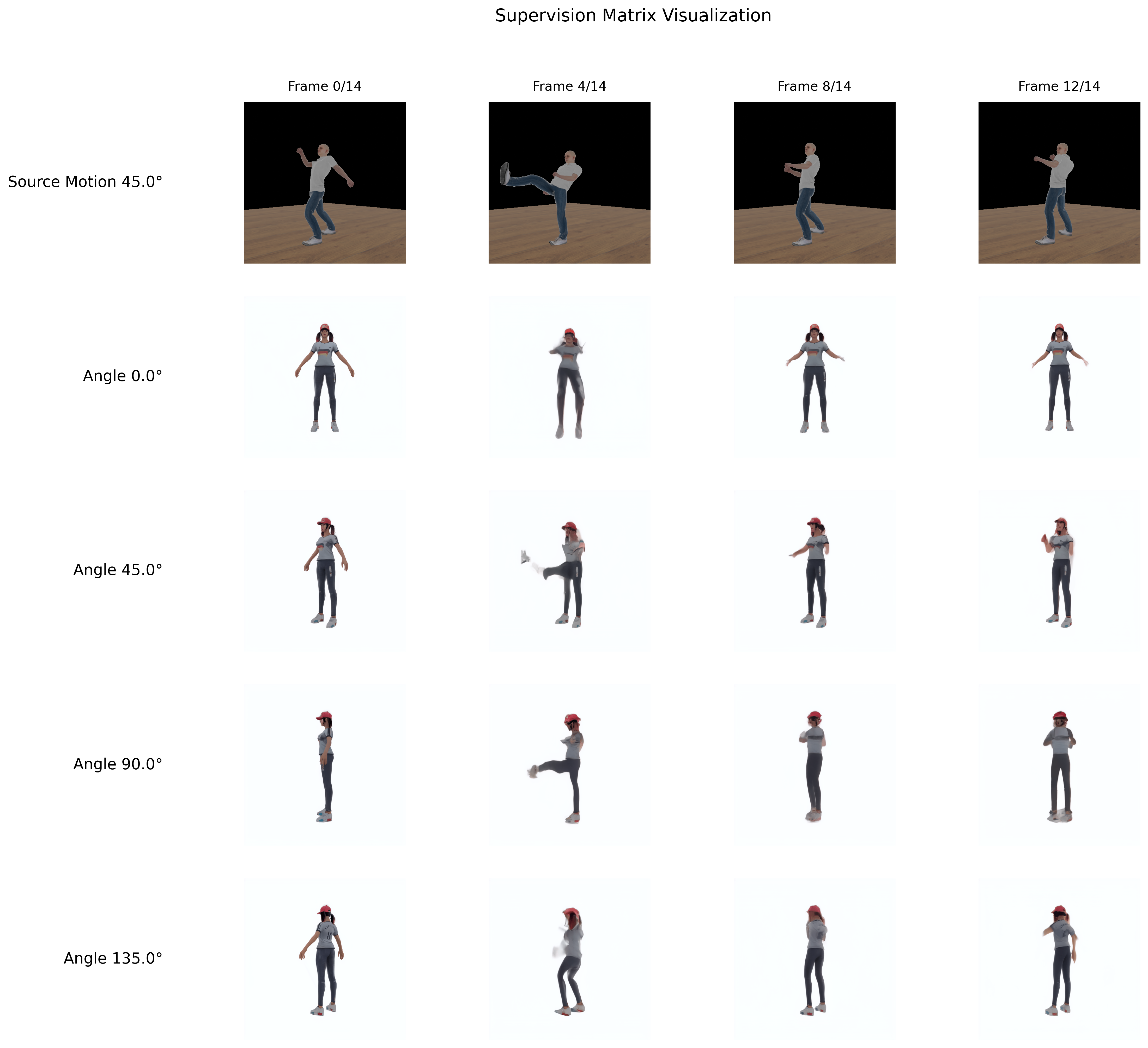}
    \caption{\textbf{Supervision videos example - Failure case.} The generated videos in this example are highly inconsistent and introduce noise into the consolidation process, resulting in visible artifacts in the final dynamic scene. }\label{fig:noisy_supervision_failure}
\end{figure}

\begin{figure}
    \centering
    \includegraphics[width=\linewidth]{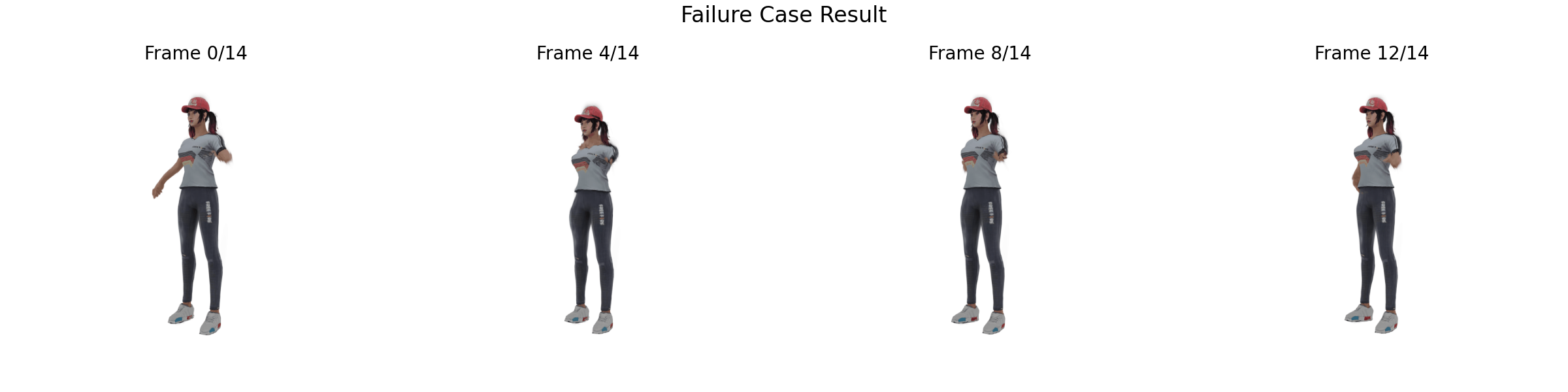}
    \caption{\textbf{Final result example - Failure case.} Final result rendered at a 45° angle, showing the kicking motion on the woman figure. The video supervisions leading to this failure are shown in Fig. \ref{fig:noisy_supervision_failure}.}
    \label{fig:final_result_failure}
\end{figure}

 Our method struggles with highly articulated motions, such as kicking or jumping jacks. In these cases, per-view motion inversion often fails, causing the synthesized reference videos to exhibit severe artifacts and lose synchronization. Once this happens, our consolidation phase cannot recover, and the artifacts propagate into the final generated motion. We illustrate this in Figure \ref{fig:noisy_supervision_failure}, which shows errors in the generated supervision for such a complex motion. Figure \ref{fig:final_result_failure} illustrates the final outcome—failure to move the legs and breakage of the thin arms. Since the main bottleneck of our method stems from the noisy video supervisions generated by the 2D video model, it is important to emphasize that our method will improve as inversion techniques in video models improve.

\section{Evaluation Metrics}
\label{app:eval_metric}

\subsection{3D Motion Fidelity Metric}
\label{app:motion_fidelity}
We observe that baselines often introduce degeneracies, such as a hand remaining static and moving along with the torso instead of articulating independently. The original Motion Fidelity metric computes a two-way trajectory distance, which fails to penalize such degenerate cases due to the lack of a bijectivity requirement. To address this, we modify the metric to measure motion fidelity in only one direction—matching target trajectories to the closest source trajectories. Formally, we define our modified Motion Fidelity (MF\textsubscript{3D}) as: 
\begin{equation} 
MF_{\text{3D}} = \frac{1}{V} \sum_{v=1}^{V} \frac{1}{m} \sum_{\tilde{\tau} \in \text{Target}} \max_{\tau \in \text{Source}} \textbf{corr}(\tau, \tilde{\tau}), 
\end{equation}

\noindent where \( V \) is the number of rendered views, \( m \) is the number of tracklets in the source view, and 
\( \text{corr}(\tau, \tilde{\tau}) \) remains unchanged:
\begin{equation}
\textbf{corr}(\tau, \tilde{\tau}) = \frac{1}{F} \sum_{k=1}^{F} 
\frac{v_t^x \cdot \tilde{v}_t^x + v_t^y \cdot \tilde{v}_t^y}
{\sqrt{(v_t^x)^2 + (v_t^y)^2} \cdot \sqrt{(\tilde{v}_t^x)^2 + (\tilde{v}_t^y)^2}}
\end{equation}

An additional advantage of this measure is that it does not require ground truth, allowing us to evaluate on both of our benchmarks.

\subsection{CLIP-I and CLIP score}
In addition to motion quality, we assess appearance preservation using \textit{CLIP-I} \cite{rahamim2024bringing,zhao2023animate124}, which measures the cosine similarity between the CLIP visual features of the initial image rendered from the static object and each frame in the predicted video. This provides a reference-free, view-invariant signal for how well the object's visual identity is preserved throughout the animation. 
We also compute \textit{CLIP similarity} between individual frames of the predicted dynamic 3DGS and reference target videos. This frame-level comparison requires both source and target to perform approximately the same motion and is therefore only applied to the Mini-Mixamo benchmark. As with Motion Fidelity, all CLIP-based metrics are averaged across rendered views. While CLIP is robust to viewpoint variation and sensitive to visual attributes\cite{jain2021putting,goh2021multimodal}, it may overestimate similarity in synthetic scenes, often producing high scores despite structural inconsistencies.

\section{Novel-view Motion Synthesis}\label{app:novel_motion_synthesis}
\begin{figure}
    \centering
    \includegraphics[width=\linewidth]{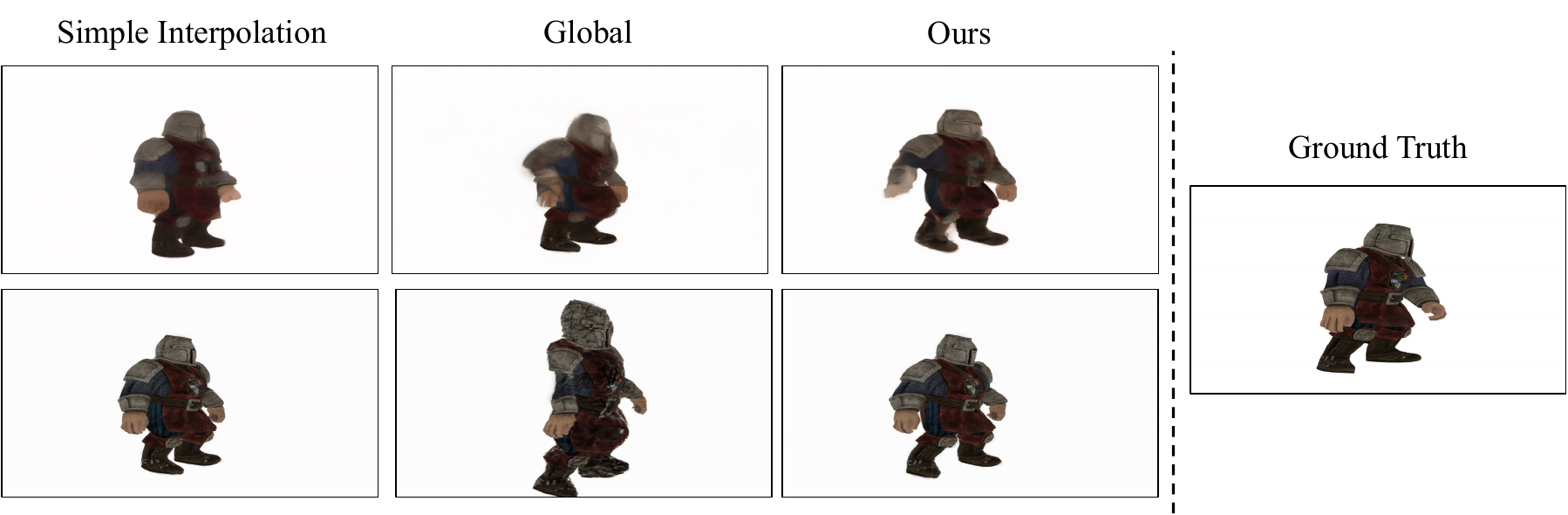}
    \caption{\textbf{Qualitative comparison of novel-view motion synthesis.} The top row shows raw supervision, while the bottom row presents results after reconstruction. Simple interpolation and global embeddings struggle to generalize to unseen views, whereas our anchor-based mechanism successfully recovers faithful motion, even for a different target.}
    \label{fig:novel_view_appendix}
\end{figure}

As explained in Section \ref{sec:novel_view}, we leverage the continuity of our anchor-based embedding representation to interpolate motion embeddings for any viewpoint between two anchors. Given a novel-view image of the source motion, we can then generate a complete video of the source object from this new perspective. In addition to Figure~\ref{fig:Novel_View} where we demonstrated the source object animated from a novel angle for which we did not obtain ground truth data, in Figure~\ref{fig:novel_view_appendix} we showcase the ability to generalize to new angles, and new objects at the same time.

\subsection{Appearance preservation}\label{app:appearance_metric}
In addition to motion quality, we assess appearance preservation using \textit{CLIP-I}\cite{rahamim2024bringing,zhao2023animate124}, which measures the cosine similarity between the CLIP visual features of the initial image rendered from the static object and each frame in the predicted video. This provides a reference-free, view-invariant signal for how well the object's visual identity is preserved throughout the animation. We also compute \textit{CLIP similarity} between individual frames of the predicted dynamic 3DGS and reference target videos. This frame-level comparison requires both source and target to perform approximately the same motion and is therefore only applied to the Mini-Mixamo benchmark. As with Motion Fidelity, all CLIP-based metrics are averaged across rendered views. While CLIP is robust to viewpoint variation and sensitive to visual attributes\cite{jain2021putting,goh2021multimodal}, it may overestimate similarity in synthetic scenes, often producing high scores despite structural inconsistencies.

\section{Additional details on Ablation}
\subsection{Inversion Ablation}
\label{app:inversion-ablation}
\begin{figure*}[ht]
    \centering
\includegraphics[width=\textwidth]{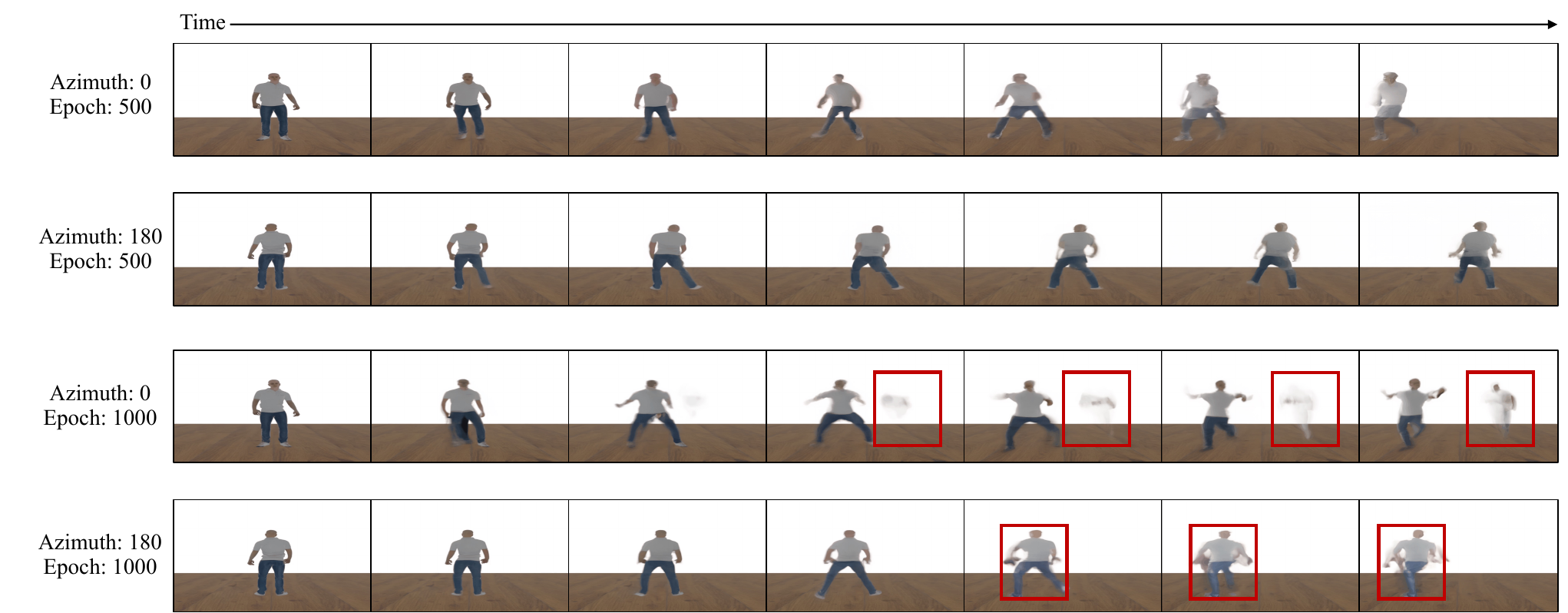}
    \caption{\textbf{Intermediate steps of the optimization process for a single motion embedding.} From an azimuth angle of 0°, the person in the video appears to be moving toward the left side of the screen, whereas from an azimuth of 180°, the 2D motion looks entirely different, with the person appearing to move toward the right. At the beginning of the training process (epoch 500, first two rows), the model starts learning the motion from the 0° viewpoint but struggles with the motion from the 180° viewpoint. By epoch 1000, the model has overfitted to the motion observed from azimuth 0°, fully learning to move the object in the first frame toward the left side of the screen. However, as a consequence, it fails to generalize to the azimuth 180° viewpoint, where the figure should move to the right but instead also moves to the left. Additionally, there is some appearance leakage from the back view, resulting in artifacts in the front view.}
    \label{fig:optimization_steps}
\end{figure*}

\paragraph{Number of anchors.} A key parameter of our method is the number of anchors, \(K\). Here, we analyze its impact on performance metrics during optimization. In Figure~\ref{fig:ablation_quant} of the main paper, each point represents the average accuracy over five random seeds, with shaded areas indicating the 95\% confidence intervals. Different
colors and markers denote the number of learnable anchors used
in training and reconstruction. Figure~\ref{fig:ablation_quant} also reveals a notable behavior: First, for both metrics, increasing the number of anchors generally decreases convergence speed. This phenomenon can be attributed to the information-sharing mechanism among anchors. When fewer anchors are used, each anchor benefits from a broader aggregation of information across multiple source views. However, this trend breaks when employing a single global embedding, which effectively acts as a single anchor. In this case, we observe the slowest convergence rate and lower final performance. This can be explained by the fact that distant views require distinct embeddings, and forcing them into a single representation results in suboptimal optimization. To support this hypothesis, we visualize this effect in Figure~\ref{fig:optimization_steps}, where motion observed from opposite viewpoints—where in one view the motion appears to the right and in another to the left—confuses a single global embedding, leading to hallucinations in the reconstructed motion. Based on this analysis, we select \( K = 5 \) anchors as a practical balance between convergence speed and reconstruction quality. 

\subsection{4D Consolidation Ablation}
\label{app:reconstruction-ablation}
\paragraph{The Effect of Regularization on 4D Consolidation Quality.}
As detailed in Section~\ref{sec:method}, we introduce several enhancements to the reconstruction stage of our pipeline, including rotational ARAP and LPIPS regularization. Figure~\ref{fig:ablation_quality} in the main paper visually demonstrates the impact of these refinements, showing significant improvements in structural integrity and fine-detail preservation. For example, text originally visible on the target remains legible even during motion (Figure~\ref{fig:ablation_quality}, + rotational ARAP column), highlighting the improved stability of our reconstructions—we encourage the reader to zoom in for details. Similarly, our refinements ensure that fine details, such as the flames on the glove, remain clearly visible throughout motion (Figure~\ref{fig:ablation_quality}, + LPIPS column), demonstrating the effectiveness of our approach in preserving high-frequency information. Importantly, applying both regularizations together further improves CLIP similarity scores, as shown in Table~\ref{tab:ablation}. Despite these clear visual enhancements, motion fidelity \textbf{scores} remain largely unchanged. This suggests that our baseline method already preserves motion effectively, but also highlights a limitation of the metric: motion fidelity is primarily sensitive to trajectory alignment rather than reconstruction quality. Even when structural inconsistencies or artifacts are present, high scores can still be achieved as long as sufficient trajectory matches exist. This gap is reflected in the \textit{human preference study} which demonsrtaes clear preference to our regularized variant in both appearance and motion preference (see Fig.~\ref{fig:human_survey}). Thus, while our refinements significantly enhance visual quality and structural consistency, these improvements are not fully reflected in motion fidelity scores.

\begin{figure}
    \centering
\includegraphics[width=\linewidth]{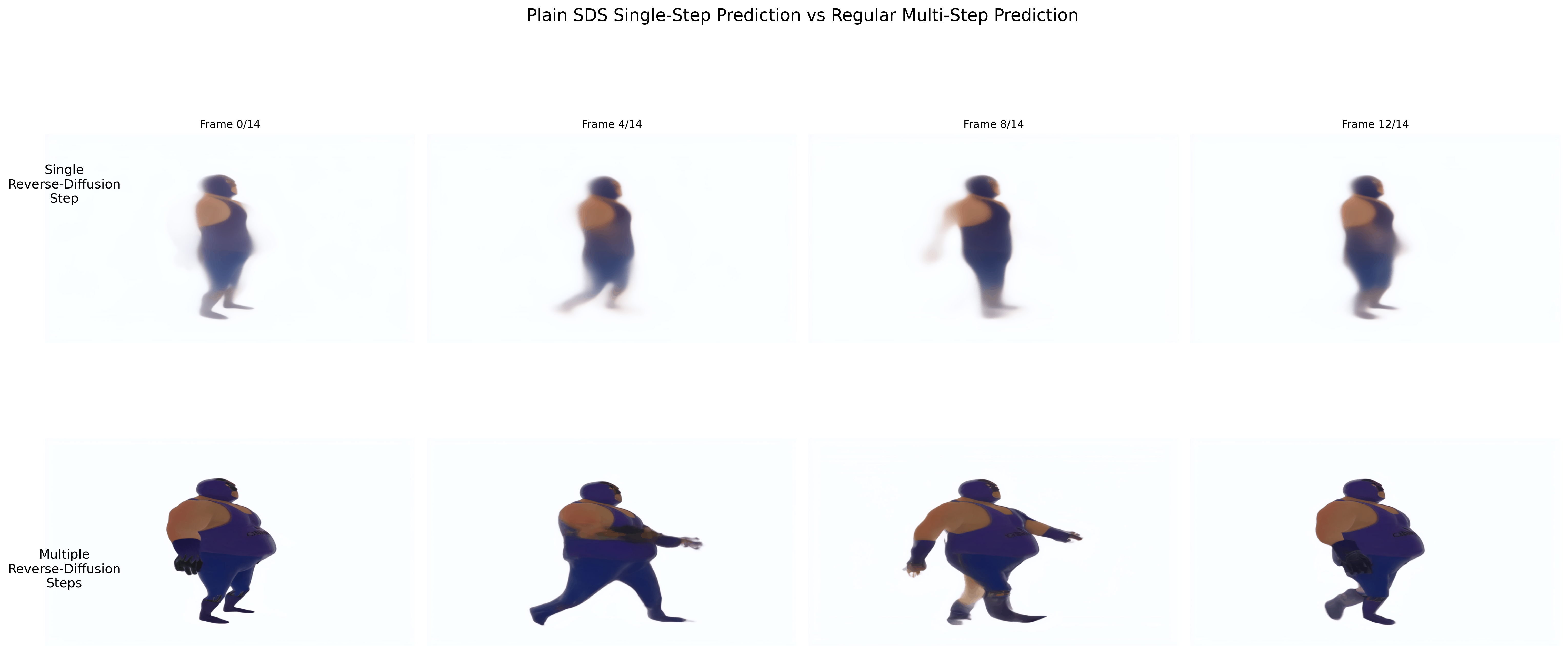}
    \caption{\textbf{SDS failure.} In our setup, when trying to use SDS-Loss for the 4D consolidation, the loss causes the Gaussians to collapse. We identify the reason as the blurry single-step denoising prediction generated by SVD. See Sec.~\ref{app:sds} for details.}\label{fig:multi_vs_single_step_DM}
\end{figure}

\textbf{SDS Loss.}\label{app:sds}
As discussed in Section~\ref{sec:method}, SDS~\cite{poole2022dreamfusion} is a widely used strategy for reconstruction with generative models under view inconsistencies, commonly adopted in 4D generation~\cite{wu2024sc4d, miao2024pla4d, zeng2024stag4dspatialtemporalanchoredgenerative, jiang2024consistentd, uzolas2024motiondreamer, bah20244dfy, bah2024tc4d, ling2024alignyourgaussians, ren2023dreamgaussian4d}. we explored SDS loss as an alternative to LPIPS loss and ARAP regularization for handling multiview inconsistencies. However, applying plain SDS in a single-step UNet inference led to blur in moving object regions (see Fig.~\ref{fig:multi_vs_single_step_DM}). This blur caused a mismatch with the sharp 3DGS primitives, prompting the optimization to overfit blurry areas with detailed splats—ultimately collapsing the shape after a few iterations.

In an attempt to address this, we experimented with the Iterative Dataset Update strategy~\cite{instructnerf2023}, which replaces single-step inference with multiple UNet steps per iteration, simulating the full generative process. It generates new supervision videos by injecting noise into renderings and denoising iteratively. However, this still produced similar artifacts, requiring ARAP regularization and LPIPS loss for stability. While we did observe slight improvement in motion quality when combined with our regularization strategy, it substantially increased optimization time (1 hour vs. 5 minutes), so we opted not to include it in our final method. Quantitative comparisons are reported in Table~\ref{tab:ablation}.

\begin{table}[h]
    \centering
    \resizebox{0.8\linewidth}{!}{ % Adjust 0.8 to a smaller value if needed
    \begin{tabular}{lcc}
        \toprule
        \textbf{Method} & \textbf{Motion Fidelity} $\uparrow$ & \textbf{CLIP Score} $\uparrow$ \\
        \midrule
        Naive & 0.7663 $\pm$ 0.0098 & 0.9423 $\pm$ 0.0019\\
        + ARAP Rotation               & 0.7567 $\pm$ 0.0108 & 0.9608 $\pm$ 0.0010 \\
        + LPIPS Loss                    & 0.7602 $\pm$ 0.0106 & 0.9636 $\pm$ 0.0010\\
        \midrule

        Naive + Iterative DU           & 0.7822 $\pm$ 0.0384 & 0.9412 $\pm$ 0.0075\\
        ARAP Rotation + LPIPS + Iterative DU                 & 0.7686 $\pm$ 0.0272 & 0.9606 $\pm$ 0.0027\\

        \bottomrule
    \end{tabular}
    }
    \caption{\textbf{Ablation study results.}}
    \label{tab:ablation}
\end{table}

\subsection{Human Preference Study}\label{app:human_study}

We include here screenshots from our survey, both from the qualitative comparison, and the ablation study.
\begin{figure}[H]
    \centering
    \includegraphics[width=0.4\linewidth]{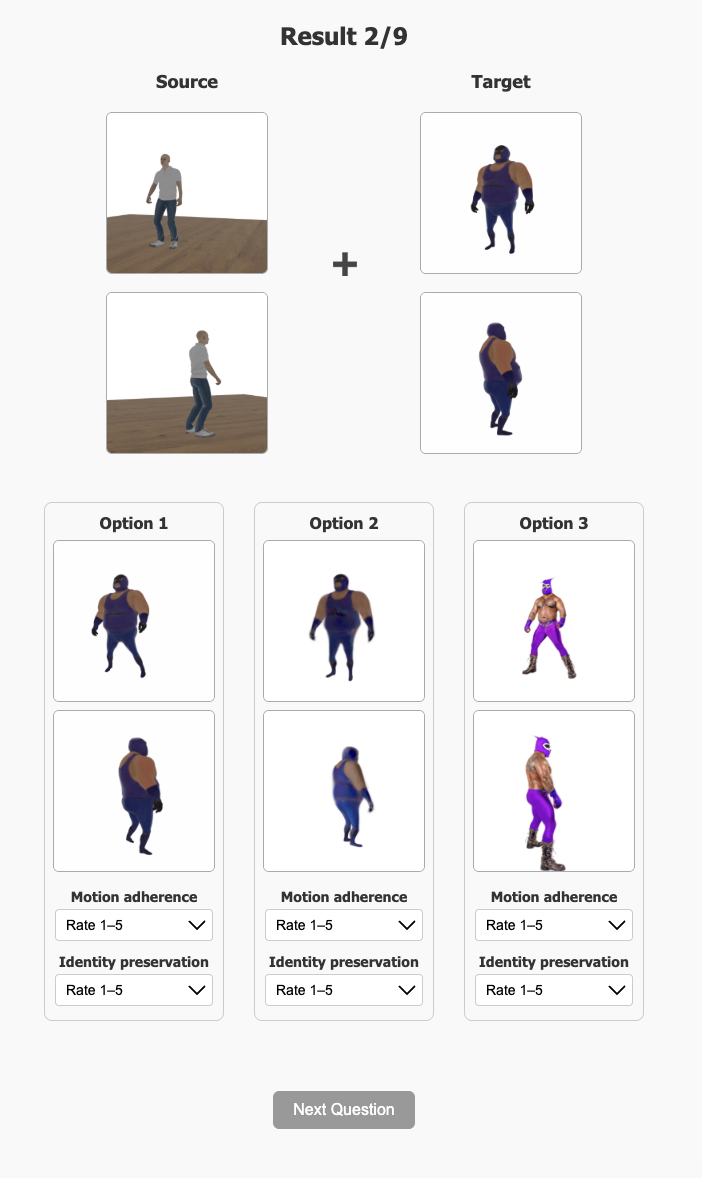}
    \caption{\textbf{Screenshot 1.} A screenshot showing the qualitative comparison survey.}
    \Description{Decorative image, no meaningful content.}
    \label{fig:survey1}
\end{figure}

\begin{figure}[H]
    \centering
    \includegraphics[width=0.6\linewidth]{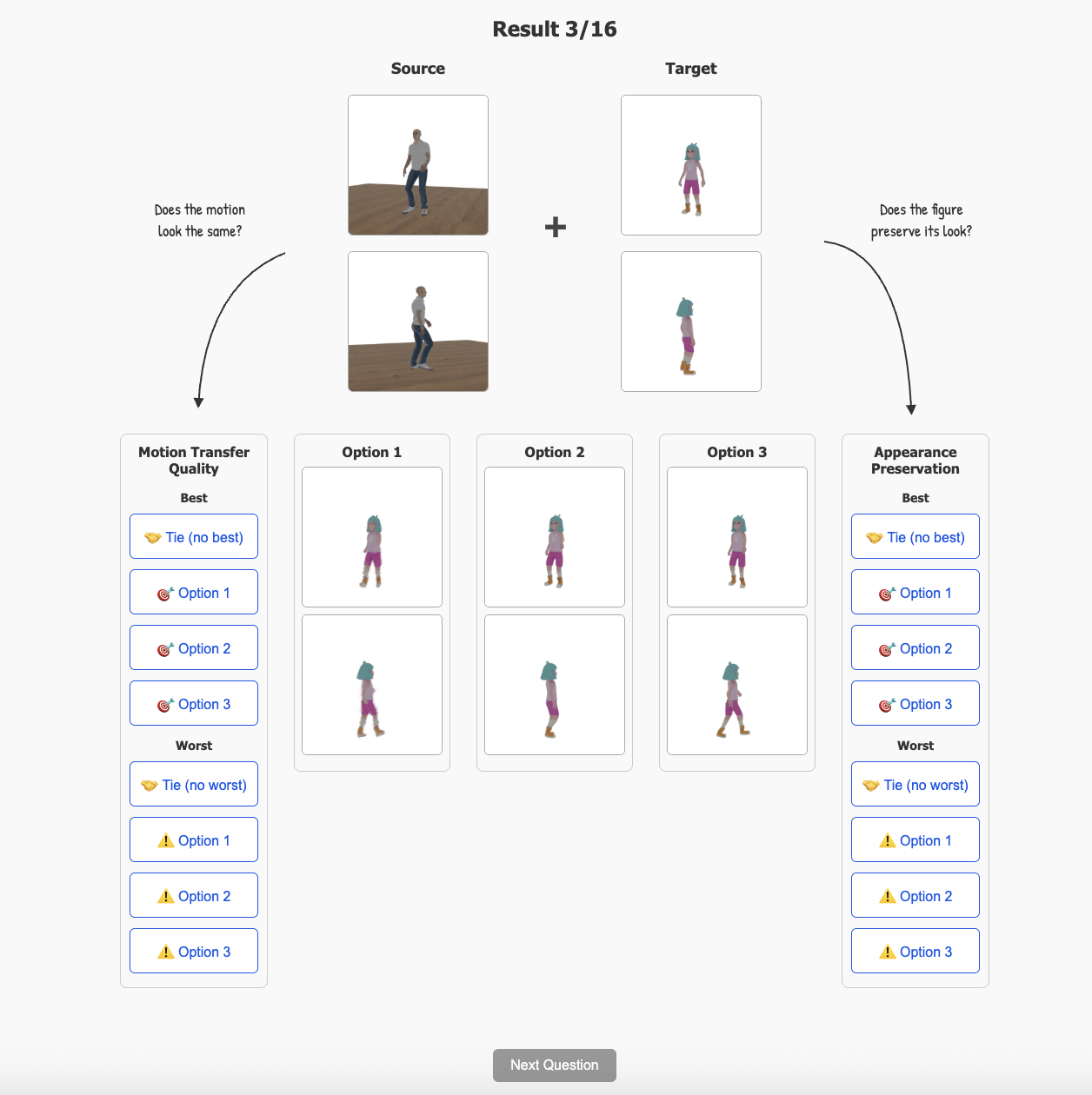}
    \caption{\textbf{Screenshot 2.} A screenshot showing the ablation user study.}
    \Description{Decorative image, no meaningful content.}
    \label{fig:survey2}
\end{figure}

% \appendix
% \section{Switching Times}

\end{document}